# Divergences in Color Perception between Deep Neural Networks and Humans

Ethan O. Nadler[a,b]*, Elise Darragh-Ford[c], Bhargav Srinivasa Desikan[d,e], Christian Conaway[f], Mark Chu[g], Tasker Hull[h], and Douglas Guilbeault[i]*

[a]Carnegie Observatories, USA; [b]Department of Physics, University of Southern California, USA; [c]Kavli Institute for Particle Astrophysics and Cosmology and Department of Physics, Stanford University, USA; [d]School of Computer and Communication Sciences, EPFL, CH; [e]Knowledge Lab, University of Chicago, USA; [f]University of California, San Diego, USA; [g]School of the Arts, Columbia University, USA; [h]Psiphon Inc., Toronto, CA; [i]Haas Business School, University of California, Berkeley, USA.

*Corresponding authors: Ethan O. Nadler (enadler@carnegiescience.edu) and Douglas Guilbeault (douglas.guilbeault@haas.berkeley.edu)

## Abstract

Deep neural networks (DNNs) are increasingly proposed as models of human vision, bolstered by their impressive performance on image classification and object recognition tasks. Yet, the extent to which DNNs capture fundamental aspects of human vision such as color perception remains unclear. Here, we develop novel experiments for evaluating the perceptual coherence of color embeddings in DNNs, and we assess how well these algorithms predict human color similarity judgments collected via an online survey. We find that state-of-the-art DNN architectures — including convolutional neural networks and vision transformers — provide color similarity judgments that strikingly diverge from human color judgments of (*i*) images with controlled color properties, (*ii*) images generated from online searches, and (*iii*) real-world images from the canonical CIFAR-10 dataset. We compare DNN performance against an interpretable and cognitively plausible model of color perception based on wavelet decomposition, inspired by foundational theories in computational neuroscience. While one deep learning model — a convolutional DNN trained on a style transfer task — captures some aspects of human color perception, our wavelet algorithm provides more coherent color embeddings that better predict human color judgments compared to all DNNs we examine. These results hold when altering the high-level visual task used to train similar DNN architectures (e.g., image classification versus image segmentation), as well as when examining the color embeddings of different layers in a given DNN architecture. These findings break new ground in the effort to analyze the perceptual representations of machine learning algorithms and to improve their ability to serve as cognitively plausible models of human vision. Implications for machine learning, human perception, and embodied cognition are discussed.

## I. Introduction

Over the last decade, deep neural networks (DNNs) have matched and even surpassed human performance on a range of visual tasks, including image classification and object recognition[1,2]. Indeed, popular accounts maintain that DNNs readily learn abstractions from noisy and raw data that are similar to the abstractions leveraged in human cognition[3,4]. A number of studies show that DNNs are effective at predicting human brain activity by modeling fMRI data associated with human vision and language processing[5–8]. These achievements have inspired arguments that DNNs offer cognitively plausible models of human visual perception and human cognition more broadly[5–7,9,10]. However, a growing body of work shows that DNNs and humans can arrive at similar image classifications (such as correctly labeling an image as containing a "dog" or a "tree") while nevertheless relying on strikingly different perceptual processes[11]. For example, a recent audit study shows that, as DNNs become more accurate in classifying images, the image features they leverage increasingly deviate from human patterns of visual attention[12]. Relatedly, "adversarial attacks" show that subtle perturbations in image input, such as randomly shuffling a small fraction of pixels, can lead to drastically different DNN classifications despite having no qualitative impact on images from a human perspective[13–15]. It remains an active debate whether DNNs perceive and represent visual information in a manner that resembles human vision or, more ambitiously, human cognition[11,16–18].

A key limitation of prior research is that it primarily evaluates the cognitive plausibility of DNNs by examining their ability to learn human categories when solving complex tasks, such as image classification and object recognition[1,3]; yet, these tasks are not designed to compare perceptual processes between DNNs and humans. Moreover, DNNs frequently develop complex and high-dimensional representations to solve such tasks. Because it is challenging to decompose and interpret these representations, it remains difficult to isolate the perceptual processes and basic image features that DNNs leverage[16,17,19]. A persistent obstacle in this domain is the challenge of creating task paradigms that allow for a clear separation of perceptual information and high-dimensional abstractions in DNNs. One recent proposal is to compare DNNs and humans in how they represent simple, validated stimuli from cognitive psychology designed to elicit and observe basic perceptual processes[19,20]. For example, recent work evaluates how DNNs represent canonical gestalt stimuli, such as schematic images exemplifying foreground-background relations, as well as visual patterns relating to object closure[11]. It is shown that DNNs often struggle to represent gestalt patterns[11], and when they do[21–23], these effects only occur in the final layers of the DNN, whereas gestalt effects can be detected in the early stages of human visual processing[24,25]. However, despite these advances, this recent work continues to examine stimuli involving a mixture of basic perceptual information (e.g., texture and color) along with more complex and abstract information (e.g., spatial and geometrical properties of images on multiple scales), which limits the ability to isolate how DNNs represent basic perceptual information as part of their more abstract representations.

We address this gap by comparing DNNs and humans in terms of how they represent a foundational aspect of human visual perception—namely, color—as presented in simple, controlled color stimuli and in the presence of more complex textural and spatial patterns. Color is particularly relevant for these explorations for two key reasons. First, color has received strikingly little attention in recent efforts to probe the "cognitive psychology" of DNNs[11,19]. Only a few recent studies examine DNN color representations — focusing on DNN embeddings of complex, real-world images — and these studies do not systematically compare against human perceptions of color[26–28]. Second, color plays a far-reaching role throughout human cognition. Color vision is ubiquitous in primates[29,30] and is known to evoke a range of behavioral, emotional, and linguistic responses in humans[31–34]. Color is frequently evoked in research on embodied cognition as a powerful example of how humans harness sensory information to support a wide range of cognitive functions, such as abstraction[35,36] and metaphorical reasoning[37–39]. The human visual system is highly tuned for color perception, and, crucially, preserves color information throughout the formation of abstractions related to visual input. It is difficult to imagine a computer vision algorithm that faithfully models human vision while failing to perceive color in a cognitively plausible manner and failing to preserve color-related information throughout the visual learning process.

Yet, there are several reasons to suspect that extant DNNs are not designed to represent color in a way that resembles human visual processing. Extant DNNs are primarily trained on images represented in RGB color space, which has been shown to fail at capturing human color perception[40,41]. Furthermore, several studies argue that DNNs are biased toward perceiving images based on the spatial arrangement of pixel luminosities—i.e., image *texture*—rather than in terms of color[42,43]. Recent attempts to characterize the representations of DNNs suggest that they may drop color information as early as their second layer, which becomes skewed toward the detection of gray-scaled geometric patterns[43,44]. DNN architectures may be prone to dropping basic perceptual information such as color in the interest of developing high-dimensional abstractions, in stark contrast to human visual cognition, which preserves color information and intermingles it with a wide array of representations, both emotional and linguistic[31,34,37,38]. Consistent with this theory, recent work finds that increasing the depth and dimensionality of DNNs, which correlates with their capacity for abstraction, harms their ability to represent color information[27]. These findings suggest that state-of-the-art DNNs for image classification may be poorly designed to capture human color perception given their bias toward abstracting away from raw sensory data and basic perceptual information. However, limitations in the ability to identify and examine the perceptual representations of DNNs have prevented prior research from demonstrating this empirically.

Here, we develop a range of novel visual experiments for evaluating the coherence of color embeddings in pre-trained DNNs, and we assess how well these algorithms' embeddings predict human color similarity judgments collected via an online survey. We provide evidence that state-of-the-art DNN architectures — including convolutional neural networks[2] and vision transformers[45,46] — do not represent color in a way that resembles or effectively predicts human color judgments of (*i*) images with controlled color properties, (*ii*) images generated from online Google Image searches, and (*iii*) real-world images from the canonical CIFAR-10 dataset[47]. These results hold when examining DNNs trained on different high-level visual tasks, including image classification and image segmentation[48] (Figure S14). All of the DNNs we examine exhibit similar shortcomings in their ability to model human color perception, with important implications for the plausibility of extant DNNs as models of human visual cognition. At the same time, we find that a convolutional DNN trained on a style transfer task (hereafter referred to as a "style transfer DNN") performs notably better than all other DNNs we test, suggesting that different training goals can influence the relevance of color in DNN embeddings[49,50]. Moreover, we replicate our analyses by comparing human color judgments against the color representations formed across all layers of a convolutional DNN trained on image classification, given recent work suggesting that earlier layers of DNNs may better represent color[27,51]. We find that, while none of the network's layers effectively predict human color perception, deeper layers actually perform better, suggesting that hierarchical learning can contribute to DNNs' abilities to represent color (Figure S13). These results shed new light on how DNN architectures and training objectives can improve their ability to represent color in a cognitively plausible manner, as discussed in detail below.

To further ground the interpretation of our DNN results, we develop an alternative algorithmic approach to modeling human color perception based on cognitively plausible wavelet transforms. Wavelet transforms comprise a simple learning architecture that is aptly positioned to efficiently detect and preserve frequency information, including that contained in color channels. Despite being developed in a scientific context separate from the study of human perception, foundational work in computational neuroscience shows that wavelet transforms capture properties of human vision, including the perception of color[52–54] and texture[25,55,56]. Wavelet transforms are well-positioned to serve as an interpretable benchmark algorithm for learning color representations against which the more complex, high-dimensional representations of DNNs can be compared. Recent studies have even found that processing images with wavelet transforms prior to DNN analysis improves classification accuracy[51,52]. We expand on these results by comparing DNN and wavelet embeddings against human color perception in an interpretable setting, thereby clarifying the cognitively relevant color features captured by wavelets and often missed by DNNs. In addition, a key limitation of many related studies is that they train DNNs in RGB color space, which is known to inaccurately model human color perception. For this reason, we implement our wavelet algorithm in both RGB color space and an approximately perceptually uniform color space that re-weights RGB channels so that Euclidean distances in color space match human perceptible differences in color[41], providing various benchmarks to compare against DNN performance.

In what follows, we demonstrate that our wavelet algorithm is significantly better than state-of-the-art image classification DNNs at predicting human color judgments across all image datasets examined, even when implemented in the same RGB color space in which these state-of-the-art DNNs are trained. This includes comparing against the style transfer DNN trained on artworks; although this network outperforms all other DNNs we test in terms of color representation, it still significantly underperforms relative to our wavelet algorithm under various conditions, regardless of whether our wavelet algorithm operates in approximately uniform or RGB color space. We further show that the color relationships detected by our algorithm are considerably more interpretable, allowing for more direct comparisons to human color vision and revealing areas of color space in which DNNs struggle to differentiate color in a perceptually coherent way. As such, these findings break new ground in our ability to algorithmically emulate human color perception, and to construct cognitively plausible models of human cognition capable of enriching ongoing theoretical research in cognitive science and artificial intelligence.

This paper is organized as follows. In **Study 1: Evaluating Algorithms' Color Embeddings**, we study how several widely-used DNNs represent color. In particular, we scrutinize how these algorithms' embeddings of images from three datasets—ranging from uniform color squares to real-world images used in classification tasks—leverage color. We benchmark these analyses by comparing algorithms' color similarity predictions to images' color similarity, calculated in an approximately perceptually uniform color space, and to our wavelet algorithm described above. In **Study 2: Comparing Color Embeddings Against Human Judgments**, we test how accurately these DNNs and our wavelet algorithm predict color similarity relative to human judgments collected via an online survey. In the **Discussion**, we summarize our results, highlight interesting areas for future work relating to machine learning, human perception, and embodied cognition.

## II. Study 1: Evaluating Algorithms' Color Embeddings

Methods

We begin by briefly describing our image datasets, computer vision algorithms, and methods for analyzing image embedding; additional details are provided in the **Supplementary Information**.

Figure 1 provides an overview of our methods: specifically, we feed each image dataset described below through three pre-trained DNNs; we then compare how these DNNs represent color and texture by clustering their embeddings, and analyzing the color distributions of images that each algorithm embeds similarly in an approximately perceptually uniform color space. We benchmark these DNN color embeddings against those obtained using our new, perceptually grounded wavelet algorithm. Example images from all of our datasets are shown in the top half of Figure 1; examples of DNN architectures and wavelet filters are shown in the bottom half of Figure 1.

**Image Datasets.** We use the following image datasets in our analyses:

- *Block and Stripe Images*: Sets of 1,000 (300 x 300)-pixel images with controlled color properties that are composed of one (block) or two (stripe) randomly-selected colors; for stripe images, colors are arranged in an alternating vertical stripe pattern. Note that the combination of two colors provides pixel-to-pixel variation, and thus a rudimentary form of image texture. Stimuli similar to our stripe images, often referred to as "visual gratings," are frequently used to investigate the core mechanisms of human visual processing[57]. All block and stripe images are presented with a colored border; our findings are robust to varying the color of the border surrounding these images (Figure S20).

- *Colorgrams*: A set of 1,000 (300 x 300)-pixel images, each of which is generated by averaging the top 100 Google Image results for a given search term. We select colorgrams from Desikan et al. (2020)[37] corresponding to the most frequently used words in English. Colorgrams visually represent concepts while heavily featuring color; in particular, previous studies show that colorgrams encode semantically meaningful relationships that effectively differentiate concrete and abstract concepts, as well as linguistic metaphors, via color[37,38].

- *CIFAR-10*: A set of 10,000 images from the standard computer vision dataset CIFAR-10. In particular, we collect 10,000 images that are commonly used to train image classification algorithms[47], consisting of 10 approximately equally-represented image classes (airplane, automobile, bird, cat, deer, dog, frog, horse, ship, and truck). For consistency with the other image datasets, we resize CIFAR-10 images from their original resolution of (32 x 32) pixels to (300 x 300) pixels each; in supplementary analyses, we show that our results are unchanged when using a higher-resolution version of CIFAR-10 (Figure S18).

**Computer Vision Algorithms.** We feed each set of images through four different computer vision approaches: (*i*) our new algorithm based on the discrete wavelet transform, (*ii*) a standard convolutional DNN trained on an image classification task[1], (*iii*) a "style transfer DNN", i.e., a convolutional DNN trained to superimpose textural motifs from images in a training class to previously unseen images[58], and (*iv*) a "vision transformer" DNN architecture trained to classify images while remaining sensitive, via an "attention" mechanism, to small-scale image features[45]. We now describe each algorithm in detail:

- *Perceptually Uniform Wavelet Algorithm*: We implement a second order wavelet transform using the Morlet wavelet family, which (like the Gabor wavelet) is known to capture aspects of human visual processing[25,52–54]. The resulting 48-dimensional wavelet embeddings capture textural properties of images on different spatial scales; for example, the "wavelet algorithm" box in Figure 1 illustrates several wavelet filters. Our wavelet implementation is released as part of our [publicly-available code](publicly-available code) (see "Data Availability" statement). We test our wavelet algorithm in both RGB color space and an approximately perceptually uniform color space, $J_zA_zB_z$, which emulates human color perception[41]. This color space is defined such that Euclidean distances between colors linearly map onto differences in human color vision, unlike standard color spaces like RGB. Crucially, in supplementary analyses, we show that the fidelity of our wavelet algorithm's color perception does not significantly degrade when operating in RGB color space. Thus, its success is not predetermined by our use of $J_zA_zB_z$ color space.

- *Convolutional DNN*: We use the ResNet model, which is trained to classify images from the [ImageNet](ImageNet) database[59]; this architecture is widely used for computer vision tasks. In our main analysis, we extract weights from the penultimate layer of the network, yielding a 512-dimensional embedding for each image; we also test different layers' embeddings in supplementary analyses. Note that the convolutional DNN we employ is *not* trained to classify images that represent abstract concepts, which often lack clear concrete physical referents. Abstract concepts are not represented in traditional image classification datasets and training objectives, and often exhibit more coherent color properties than images of concrete objects[38]. Nonetheless, image classification algorithms trained on ImageNet and similar datasets are widely used and have been claimed to capture aspects of human visual processing[3,7], particularly of concrete objects, making this convolutional DNN a useful test case for our study.

- *Style Transfer DNN*: We test a "style transfer" convolutional DNN that has been trained to represent the artistic style of images and to superimpose this style on previously unseen images. We use the style transfer algorithm described in Ghiasi et al. (2017)[58] by extracting 4096-dimensional image embeddings from the VGG19 network's penultimate layer. Style transfer algorithms yield images of high perceptual quality, for example using paintings from well-known artists[49,50]. As a result, the style transfer DNN provides a useful point of comparison for our study because it is trained to capture textural and color-based information in images that may not be leveraged for traditional image classification tasks. We will show that the style transfer DNN captures some aspects of human color perception better than all other DNNs we examine.

- *Vision Transformer DNN*: Finally, we test a recent vision transformer DNN architecture trained on image classification. Inspired by advances in NLP that use attention mechanisms for language modeling[60], the vision transformer was conceived as a way to substitute convolutions in CNNs with attention heads to capture local associations in images[45]. Vision transformers require more data and parameter tuning than convolutional DNNs, but often perform as well or better on image classification and object recognition tasks[15,61]. Furthermore, state-of-the-art image classification processing frameworks often combine transformer and convolutional layers[62,63]. We extract 196-dimensional embeddings from the final layer of the algorithm from Dosovitskiy et al. (2021)[45]. Although vision transformers require more data than convolutional DNNs to train from scratch, a major appeal of such algorithms is their capacity for pre-training, justifying our use of a pre-trained model.

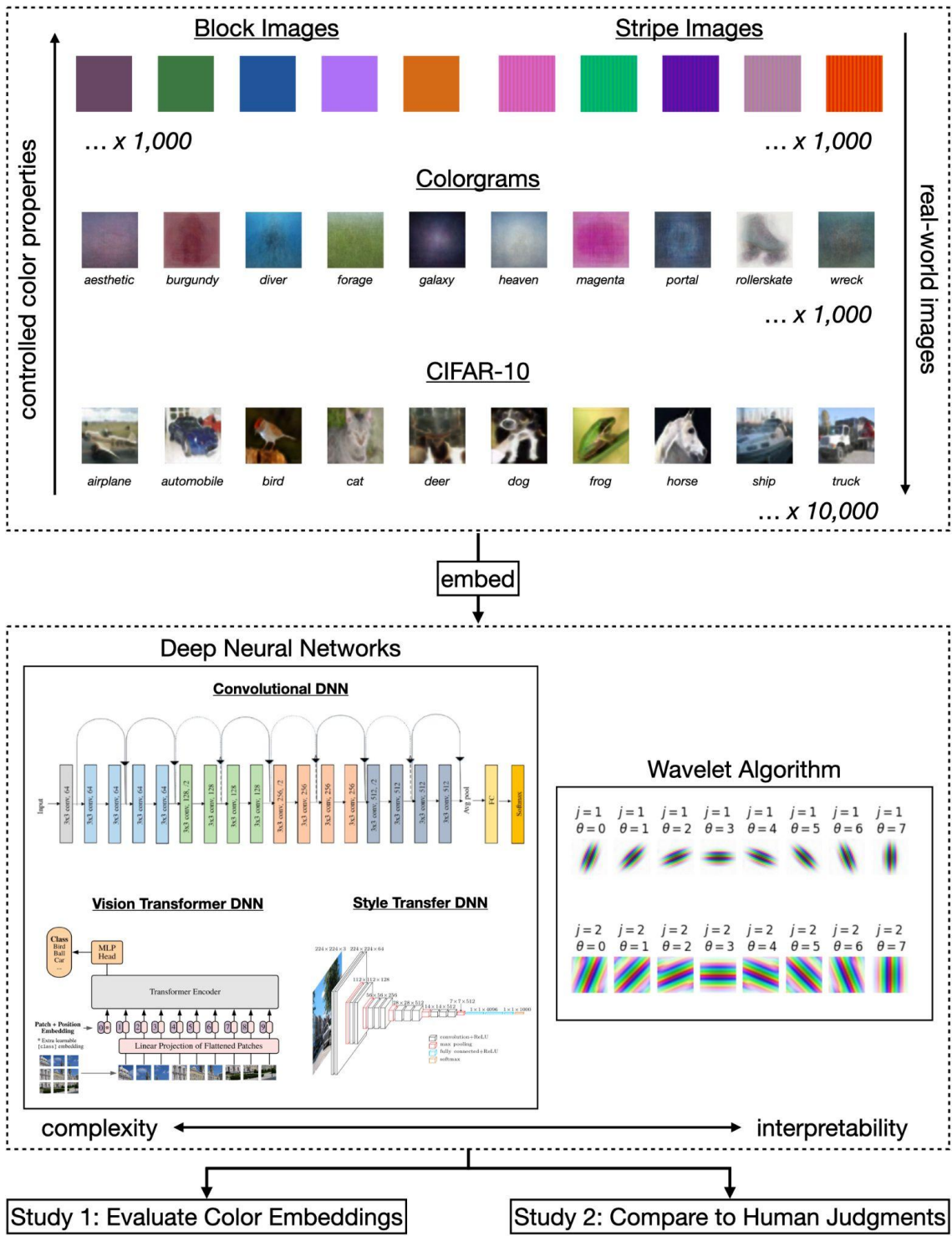

**Figure 1.** Overview of our methods for evaluating computer vision algorithms' color embeddings. DNN architectures are reproduced from Ramzan et al. (2019; convolutional DNN architecture trained on image classification)[64], Simonyan & Zisserman (2015; VGG convolutional DNN architecture trained on a style transfer task)[65], Dosovitsky et al. (2021; vision transformer DNN architecture

trained on image classification)[45]. Wavelet illustrations are reproduced from Kymatio (Andreux et al. 2018)[66].

**Analysis Techniques.** We study each algorithm's representations of the images in the datasets described above by clustering images based on their embeddings. We then analyze the color distributions of the resulting image clusters in perceptually uniform color space as described below; additional details are provided in the **Supplementary Information**.

- *Clustering Techniques***:** For each image dataset, we aggregate the embeddings returned by each computer vision algorithm. For the wavelet algorithm, we use a vector of the 16 coefficients from each perceptually uniform color channel as described above, yielding a 48-dimensional representation. For the convolutional, style transfer, and vision transformer DNNs, we use vectors of the weights from the penultimate layer of each network, yielding 512, 4096, and 196-dimensional representations, respectively. We group images according to each algorithm's embeddings using *k*-means clustering, with a dimensionality reduction factor that yields approximately ten clusters for each image dataset; this choice yields a sufficient number of images per cluster to evaluate within-cluster color distribution statistics and a sufficient number of clusters to evaluate overall differences among the algorithms' embeddings. All results are robust to alternative techniques for implementing this *k*-means clustering algorithm, including the k-init++ initialization algorithm (used in our main results, below) and a random initialization of *k*-means clusters (Figure S15).

- *Color Coherence Measure***:** We measure the color coherence of the image groups returned by *k*-means clustering by measuring the distribution of $J_zA_zB_z$[41] values, concatenated over pixel coordinates, for each image. We quantify the similarity between the color distributions of pairs of images using the Jensen-Shannon divergence and $c_{ij} = (c_i + c_j)/2$; this metric is related to the mutual information shared by two color distributions and represents a distance measure between distributions in perceptually uniform color space. In the analyses below, calculate the color similarity between all image pairs within each cluster; we also compute the mean color similarity within each cluster to study how image properties vary among clusters. We emphasize that our color coherence metric is directly based on distances in $J_zA_zB_z$ color space, which has been extensively calibrated to match human color perception[41]. Thus, our measure captures perceptual differences in color similarity and is not arbitrary.

Results

**Visual Properties of Image Clusters.** Our wavelet algorithm yields clusters that contain images with significantly more similar color distributions than state-of-the-art DNNs. Figure 2 illustrates this by comparing images from clusters returned by our perceptually uniform wavelet algorithm and the three DNNs we test for our stripe, colorgram, and CIFAR-10 datasets. The wavelet algorithm groups images with visually similar color distributions, while the DNN image classification algorithms often group images with noticeably different color properties. Similar trends hold for all image clusters and are not specific to the examples shown in Figure 2; we provide all image clusters in an online repository. We note that CIFAR-10 images often exhibit less visually coherent color distributions than the other datasets, as expected for images mainly composed of real-world objects and scenes.

**Clustering in Perceptually Uniform Color Space.** We now quantify the perceptually uniform color similarity of the image clusters returned by each algorithm. Figure 3 shows the convex hull of each stripe image cluster returned by our perceptually uniform wavelet and convolutional, style transfer, and vision transformer DNN algorithms, where each image is represented by its average coordinate in $J_zA_zB_z$ color space. It is visually apparent that the DNNs' clusters overlap in color space while our wavelet algorithm separates images with distinct color properties. The style transfer DNN's color space overlap is particularly noticeable, which may be caused by the trivial textural properties of the stripe image dataset.

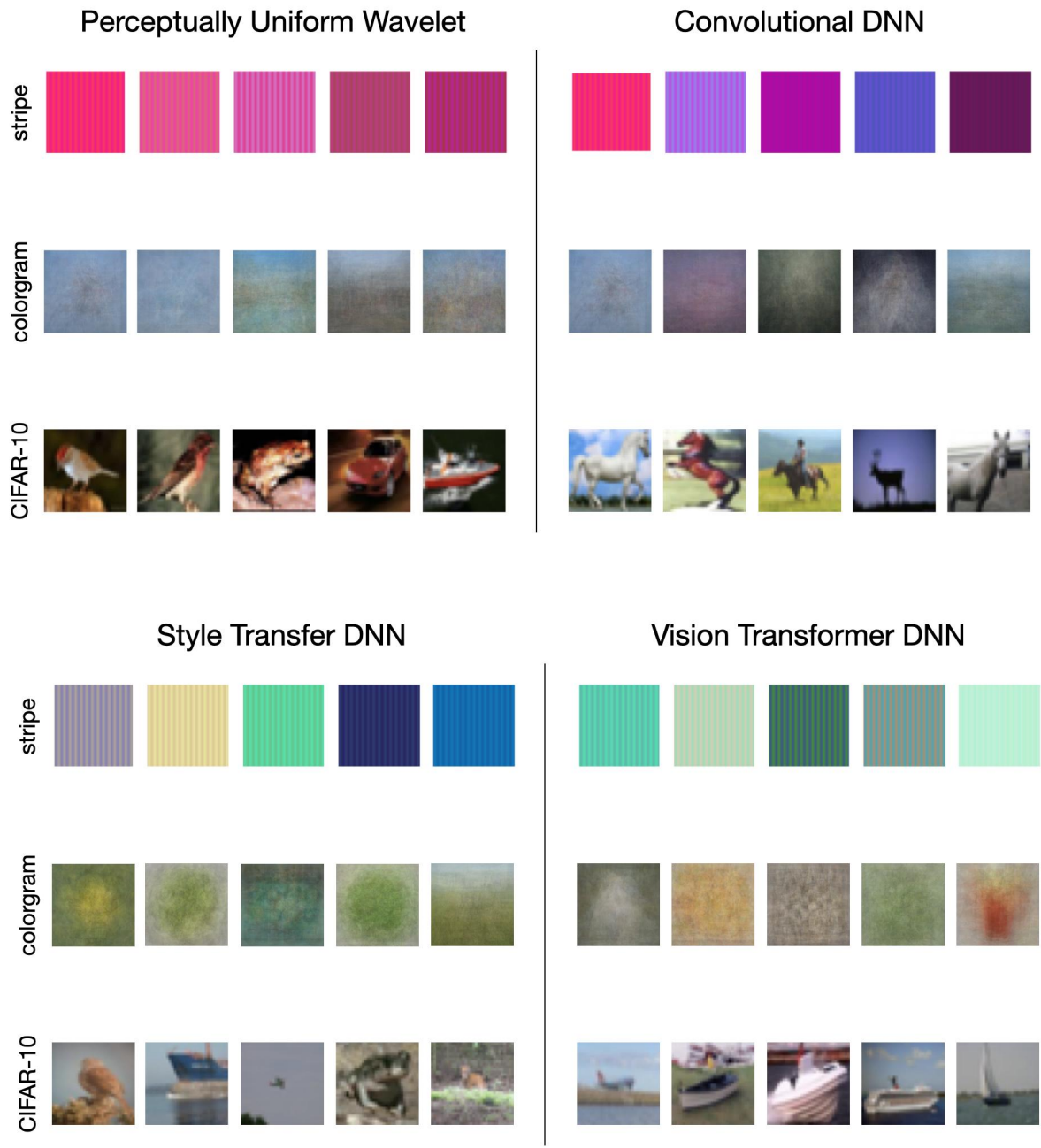

**Figure 2.** Examples of images clustered by our wavelet algorithm that operates in an approximately perceptually uniform color space (top left), by widely-used convolutional DNNs trained on image classification (top right) and style transfer (bottom left), and by a vision transformer DNN architecture trained on image classification (bottom right). For each algorithm, the top, middle, and bottom rows respectively show five images within a single cluster of similar embeddings, for our stripe, colorgram, and CIFAR-10 image datasets, respectively. Our wavelet algorithm groups images that exhibit noticeably more similar color distributions than those returned by the DNNs, on image datasets with both idealized and real-world color properties. All algorithms' image clusters are available in an online repository.

To quantify these color clustering overlap trends, we define the color coherence fraction, $f$, as the number of images that lie within the convex hull of exactly one color cluster divided by the total number of images. Thus, $f = 1$ corresponds to a maximally color-coherent grouping in which each cluster occupies a distinct region of color space, and lower values of $f$ correspond to groupings with less similar color distributions within each cluster. Note that our color clustering measure is related to the Jaccard index.

Our wavelet, convolutional DNN, style transfer, and vision transformer algorithms yield $f_{wavelet}$ = 0.81, 0.92, 0.10, $f_{CNN}$ = 0.22, 0.10, 0.04, $f_{style}$ = 0.11, 0.48, 0.19, and $f_{transformer}$ = 0.41, 0.09, 0.01 on our stripe, colorgram, and CIFAR-10 datasets, respectively. Thus, for image datasets with idealized color properties, our wavelet algorithm groups images in distinct regions of perceptually uniform color space much more effectively than any of the DNNs we study. We quantify the significance of this result by generating 100 realizations of random image clustering, which return $f_{rand}$ = 0.11 ± 0.005, 0.06 ± 0.004, 0.05 ± 0.004 for the stripe, colorgram, and CIFAR-10 datasets respectively. The strength and typical variability among these random clustering assignments is very small compared to $f_{wavelet}$, indicating that our wavelet's color clustering signal is highly significant. Note that, although the style transfer algorithm yields the most color-coherent clustering for CIFAR-10, its performance on the stripe and colorgram datasets is statistically consistent with a random clustering algorithm that is unable to detect color.

To quantitatively compare the results of our wavelet algorithm that operates in $J_zA_zB_z$ color space with a version of this algorithm that operates on images represented in RGB color space, we recompute color coherence fractions in the RGB case for all image datasets, finding $f_{wavelet,RGB}$ = 0.76, 0.84, 0.10 on our stripe, colorgram, and CIFAR-10 datasets, respectively. These results are comparable to (though slightly less color-coherent than) those returned by our fiducial wavelet algorithm, indicating that its success is not mainly driven by the fact that it represents images in an approximately perceptually uniform color space. In supplementary analyses, we provide direct visual comparisons of the image clusters returned by each version of the algorithm, confirming this finding (Figure S12).

**Color Coherence of Image Clusters.** Next, we examine the distribution of color similarity between image pairs in each cluster for each algorithm and dataset. Figure 4 shows an example of this distribution for our colorgram dataset; in particular, the filled blue histogram shows the perceptually uniform wavelet result, and the unfilled black, green, and cyan histograms show the convolutional, style transfer, and vision transformer DNN results. Our wavelet algorithm yields significantly higher pairwise color similarity, indicating that it embeds images with similar perceptually uniform color distributions more closely than the other algorithms we consider.

The unfilled blue histogram in Figure 4 shows our wavelet algorithm run in RGB color space, which still performs extremely well, indicating that our fiducial algorithm's color perception success is not simply a result of the input color space, but is instead mainly due to its architecture and mechanics. Supplementary analyses replicate these results for the stripe and CIFAR-10 datasets (Figure S16). Another indication that the mechanics of our wavelet algorithm — rather than the color space it operates in — underlie its success is that the wavelet algorithm still performs strikingly well when analyzing grayscale images (see Figures 4 and S16). While operating in grayscale significantly degrades the performance of the wavelet algorithm, as expected, Figure 4 shows that this grayscale version of our algorithm still manages to outperform the vision transformer DNN (with access to color) in terms of color clustering for the colorgram dataset, and Figure S16 shows that the grayscale wavelet still outperforms the style transfer DNN (with access to color) when clustering images by color in the stripe dataset. These findings suggest that the wavelet algorithm is able to capture information about pixel luminosity in a way that correlates with images' color distributions, whereas some DNNs with access to both pixel luminosity and color information attend to image features that do not correlate with images' overall color distributions.

We quantify the significance of color clustering results by measuring the pairwise color similarity distribution for 100 realizations of random clustering assignments; the filled vertical band in Figure 4 indicates the resulting 95% confidence interval for the mean of each realization. Strikingly, the style transfer and convolutional DNNs yield mean color similarity values that are statistically consistent with random clustering assignments that are unable to detect color by design. Furthermore, the difference between the mean of the wavelet and convolutional DNN distributions is highly significant relative to this expected spread ($p < 0.01$, Student T-test, Two-tailed). This finding also holds for CIFAR-10 images (*SOM*). As shown in Figure 4, the mean pairwise color similarity of the convolutional, style transfer, and vision transformer DNNs is consistent with our grayscaled wavelet algorithm at a level well within the statistical variation of random clustering assignments. Thus, the DNNs we test yield image clusters with color properties that are consistent with a random clustering algorithm that cannot detect color.

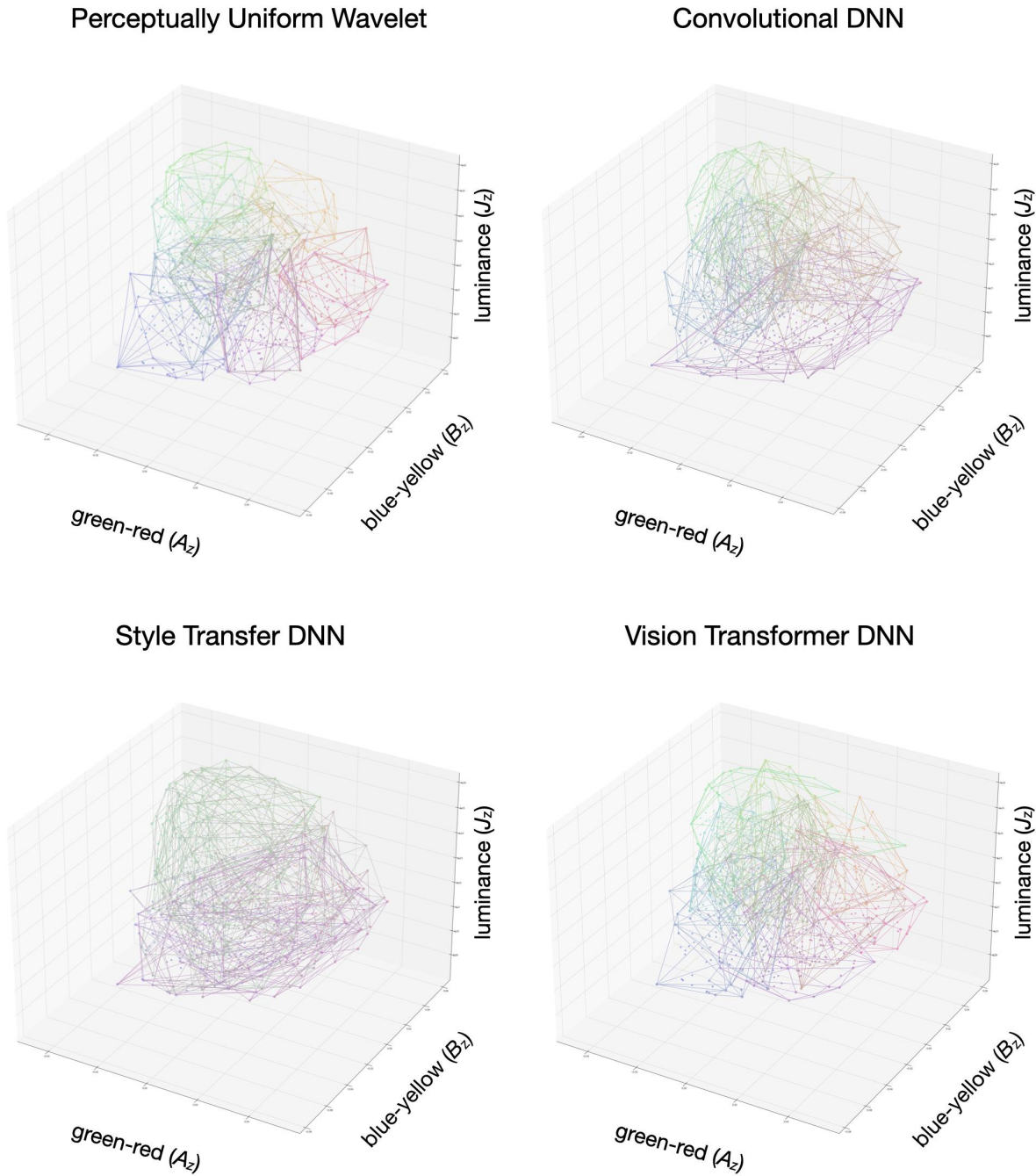

**Figure 3.** Color properties of images clustered by our wavelet algorithm (top left), by convolutional DNNs trained on image classification (top right) and style transfer (bottom left), and by a vision transformer DNN trained on image classification (bottom right). Each point shows the perceptually uniform color of a stripe image. Vertices enclose images clustered by each algorithm and are colored by clusters' mean $J_zA_zB_z$ coordinates. Our wavelet algorithm clusters images in distinct regions of perceptually uniform color space, while the DNNs mix color properties. Our wavelet algorithm sorts 81% of the stripe images into a unique cluster in color space, compared to 22%, 11%, and 41% for the convolutional, style transfer, and vision transformer DNNs.

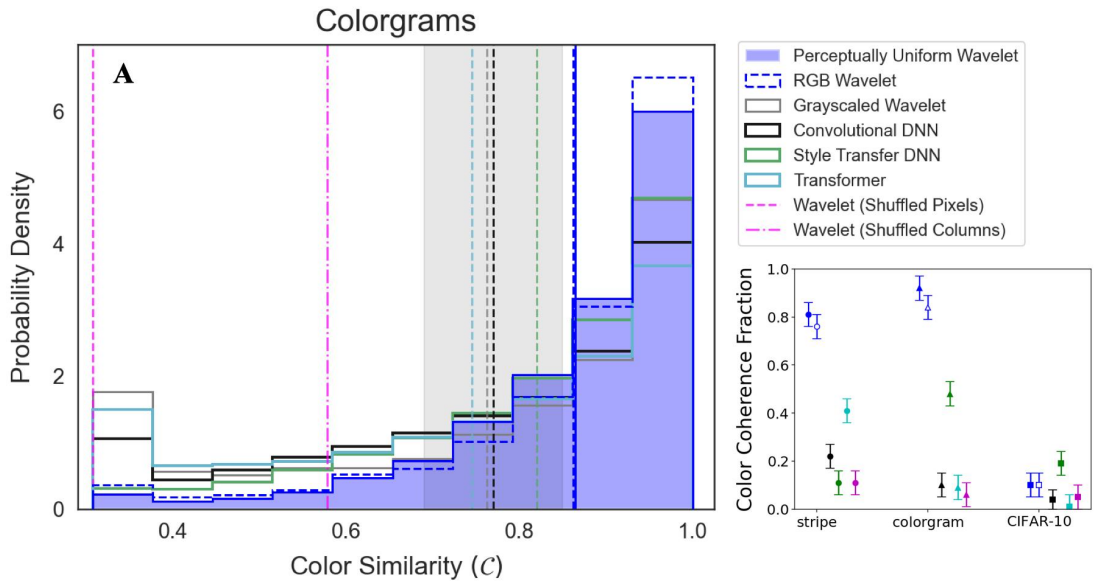

**Figure 4. A.** Comparison of the color similarity of images clustered by our wavelet algorithm and widely-used DNNs for our colorgram dataset. Color similarity distributions for all image pairs within clusters returned by our wavelet algorithm operating in perceptually uniform (blue), RGB (unfilled blue), and grayscaled (gray) color spaces, and by convolutional (black), style transfer (green), and vision transformer (cyan) DNNs. The dashed vertical lines indicate the mean of each distribution, and the filled vertical band shows the 95% confidence interval of the color similarity mean for a random clustering algorithm that does not detect color. Our perceptually uniform wavelet algorithm returns significantly more color-coherent clusters than any DNN we analyze. Furthermore, the DNNs group images with color similarity statistics that are consistent with those from a random clustering algorithm that is unable to detect color. Vertical magenta lines show the mean color similarity for the same wavelet algorithm operating on images with spatially randomized pixels, either shuffled with any other pixel in each image (dashed) or within each column of each image (dot-dashed). **B.** The fraction of images that occupy similar regions of perceptually uniform color space and that are embedded similarly by our perceptually uniform wavelet algorithm (blue), the same algorithm operated on images in RGB color space (unfilled blue), a spatially randomized version of this algorithm (magenta), by convolutional DNNs trained on image classification (black) and style transfer (green), and by a vision transformer DNN trained on image classification (cyan).

Note that our wavelet algorithm's success is not solely due to its treatment of color. In particular, when pixels within images are spatially randomized, our wavelet algorithm's color clustering performance significantly degrades relative to our fiducial results, implying that the spatial and color information captured by the wavelet transforms are correlated (see the "shuffled" results in Figure 4, magenta lines).

Specifically, we compare the perceptually uniform color similarity and embedding similarity for all image pairs in our block dataset, where color similarity is defined in Equation 1 and embedding similarity is given by the cosine similarity of image embeddings for each algorithm. We expect algorithms with more accurate color perception to exhibit a stronger correlation between the two similarity metrics. Figure 5 shows the results of this test for our perceptually uniform wavelet and DNN algorithms, confirming this expectation: the Spearman rank correlation coefficient between color and embedding similarity is 0.95 for our wavelet algorithm, versus only 0.5 for the convolutional DNN. Moreover, the convolutional DNN displays much more scatter in embedding similarity at fixed color similarity than our wavelet algorithm.

These qualitative results also hold for the style transfer and vision transformer DNNs. In particular, the style transfer DNN embeds block images similarly to the convolutional DNN, with somewhat less scatter (Spearman $\rho = 0.75$), and the vision transformer DNN differentiates block pairs of different colors most weakly (Spearman $\rho = 0.42$). For both algorithms, the overall shape and scatter of the embedding versus color similarity relation is similar to the convolutional DNN. The vision transformer DNN's relatively poor performance on this perceptual task is interesting, since it performed best, among the DNNs, at clustering stripe images by color in a perceptually coherent manner (Figure 4B). Thus, the vision transformer may require non-trivial texture—in the form of the pixel-to-pixel variation provided by our stripe images but not by our block images—to generate meaningful embeddings.

This test allows us to identify color relationships that contribute to a given algorithm's lack of perceptual color representation. For example, the top-left quadrant of both panels in Figure 5 illustrates that the convolutional DNN embeds red, blue, and green images too similarly compared to their distance in perceptually uniform color space, which is surprising given that this algorithm is trained in RGB color space. Meanwhile, the bottom-right quadrant suggests that green and yellow hues are embedded less similarly than their distance in perceptually uniform color space warrants. Our wavelet algorithm tends to embed image pairs slightly more similarly than their distance in color space warrants; however, this effect is not strongly dependent on the colors of images in each pair, implying that a constant offset in embedding similarity may further improve our wavelet algorithm's color embeddings.

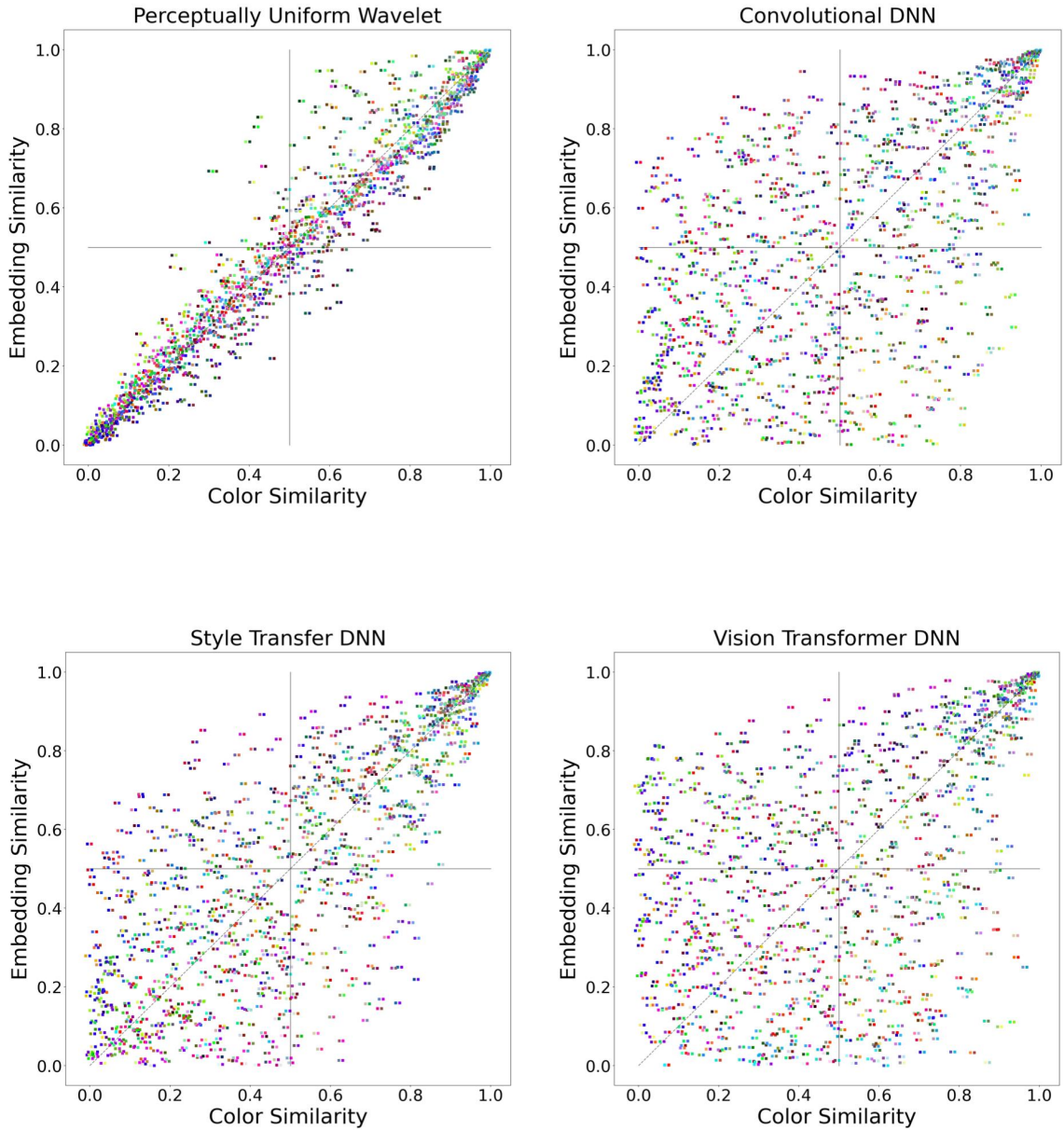

**Figure 5.** Embedding similarity versus color similarity for a random sample of image pairs from our color block dataset. Embedding similarities are shown for our perceptually uniform wavelet algorithm (top left), by widely-used convolutional DNNs trained on image classification (top right) and style transfer (bottom left), and by a vision transformer DNN trained on image classification (bottom right). In each panel, embedding and color similarities are calculated between each pair of neighboring squares, ranked from least to most similar, and minmax-normalized; dashed lines show one-to-one relations. Our wavelet algorithm's embedding similarities correlate more strongly with color similarity than for any of the DNNs (Spearman ρ = 0.95 for the wavelet algorithm, versus ρ = 0.5, 0.75, and 0.42 for the convolutional, style transfer, and vision transformer DNNs, respectively).

### III. Study 2: Comparing Color Embeddings against Human Judgments

The results above indicate that direct comparisons of algorithmic and human color perception are crucial to understand DNNs' perceptual inconsistencies when representing color. Here, we conduct such a test via an online survey.

Methods

**Experimental Design.** We designed an online experimental task in which participants compare and rank the color similarity of block image pairs. We mainly tested image pairs for which the similarity derived from our wavelet and convolution DNN embeddings were most different; we also included several "benchmark" pairs for which the algorithms were in good color similarity agreement. Following prior work, we designed our survey as a ranked grouping task to circumvent the need for quantitative, absolute color similarity judgments (see Figure 6)[67]. Details of the survey were as follows:

- *Stimuli Generation*: We tested 140 block image pairs for which our perceptually uniform wavelet and convolutional DNN algorithms' embedding similarity strongly disagreed. We also selected 60 "benchmark" pairs with comparable convolutional DNN and wavelet embedding similarities for a total of 200 color tile pairs. We created 200 "sets" of these color tile pairs by randomly assigning each pair to another, without replacement, yielding 96, 88, and 16 sets with zero, one, and two benchmark pairs, respectively.

- *Experiment & Participants*: The survey was implemented as a web-based online experiment. We surveyed 100 participants with self-reported normal color vision. Each participant was presented 25 comparison sets of our color tile pairings at random; each comparison set contained two color blocks, where each color block contained two equally sized color tiles, both constituting half the size of the block. Each participant was then asked, for each comparison set, to place the color block containing the more similar color tiles into a designated box. Each comparison set was evaluated by 12 participants on average. The survey took participants an average of 4.4 minutes to complete. See Figure 6 for screenshots of this task and its instructions as experienced by participants.

- *Outcome Measure*: For each comparison set, we identify the majority human judgment of which color block contained the most similar color tiles. Then, for the same comparison sets, we identify which block is associated with the most similar color tiles according to the embeddings of each computer vision algorithm. For our main outcome measure, we calculate the fraction of comparison sets in which each algorithm's selection of the most similar color block matches those similarity judgments made by the majority of survey respondents.

Results

Considering all 200 sets of color tile pairs, our perceptually uniform wavelet algorithm outperforms DNNs at predicting color similarities that match the judgments of color similarity provided by the majority of human participants. As shown in Figure 7, two of the DNNs fail to predict majority human color judgments significantly better than random, namely the vision transformer (46.5% accurate, $p = 0.36$, Proportion Test, Two-tailed) and the convolutional DNN (56.5% accurate, $p = 0.08$, Proportion Test, Two-tailed). The style transfer DNN is able to predict majority human color judgments better than chance (66% accurate, $p < 0.001$, Proportion Test, Two-tailed), but is not significantly better than the CNN ($p = 0.06$, Proportion Test, Two-tailed). Meanwhile, our perceptually uniform wavelet algorithm is significantly more accurate than the vision transformer and convolutional DNNs. Specifically, our wavelet algorithm is 24.5 percentage points more accurate than the transformer model ($p < 0.001$, Proportion Test, Two-tailed) and 14.5 percentage points more accurate than the convolutional DNN ($p < 0.001$, Proportion Test, Two-tailed), while the 5 percentage point difference between wavelet and style transfer DNN accuracy on this task is not statistically significant ($p = 0.33$, Proportion Test, Two-tailed).

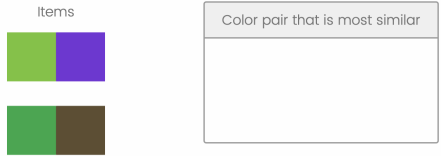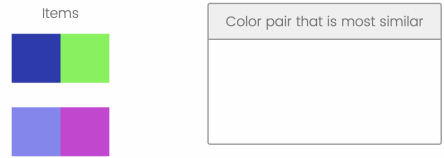

**Figure 6.** Examples of the color similarity task completed by our human respondents.

Interestingly, the embeddings of our wavelet algorithm track human judgments nearly as well as a model of perceptually uniform color space itself; its overall agreement with majority human judgments is 71.0%, compared to 73.0% when color similarities are predicted directly based on distances in $J_zA_zB_z$ color space. This success is *not* simply due to the fact that our wavelet algorithm operates in an approximately perceptually uniform color space; an alternative version of the algorithm using RGB channels performs almost equally well, yielding a 69.0% agreement rate. Together with our discussion related to Figures 4 and S16 above, this confirms that our wavelet results are robust across different input color spaces.

For the 96 sets of color tile pairs that do not contain a benchmark pair (where a benchmark pair is a pair for which our wavelet and convolutional DNN embeddings agree in terms of color similarity), the outsized performance of our perceptually uniform wavelet algorithm is particularly striking. None of the DNNs are able to successfully predict majority human judgments of the non-benchmark pairs above random chance. The style transfer algorithm only matches 54.1% of majority human judgments for these pairs, which is indistinguishable from random ($p = 0.55$, Proportion Test). The CNN (vision transform) performs even more poorly, only predicting 40.6% (38.5%) of majority human judgments for non-benchmark pairs, which is significantly worse than random ($p < 0.001$, Proportion Test, Two-tailed, for both algorithms). By contrast, our perceptually uniform wavelet algorithm is able to successfully predict majority human judgments of non-benchmark pairs with 66.6% accuracy, which is significantly higher than random ($p<0.01$, Proportion Test, Two-tailed). For the 104 color sets with at least one benchmark pair, the algorithms predict 75.0%, 76.9%, 71.1%, and 53.8% of majority human judgments, for the style transfer, wavelet, CNN, and vision transformer algorithm respectively.

Yet, comparing against the majority human judgment for each color comparison is a conservative test, since it does not leverage the rate of agreement across participants for each color comparison. Importantly, our results are robust—and in fact, are even stronger—if we compare algorithms' ability to match all individual judgments collected in our survey. We show this explicitly in the right panel of Figure 7, which shows that the wavelet algorithm matches 62% of color similarity judgments across all individuals, which is significantly higher than all DNNs. In particular, it is significantly higher than the vision transformer, which only matches 46% of individual judgments ($p < 0.001$), the CNN, which only matches 43.5% of individual judgments ($p < 0.001$), and the style transfer algorithm, which only matches 56% of individual judgments ($p < 0.001$), (Proportion Test, Two-tailed).

We also examine whether each algorithms' color similarity judgments can predict the level of agreement among separate human annotators' color judgments in the survey data. When agreement among separate human annotators is especially high, algorithms' color embeddings should reflect comparably strong measures of color similarity and difference. For each set of color tile pairs that do not contain a benchmark pair, we calculate the difference in embedding similarity of the pair that the majority of respondents deemed "more similar" versus the minority pair. We compare this to the percentage of human voters who responded that the majority pair is more similar.

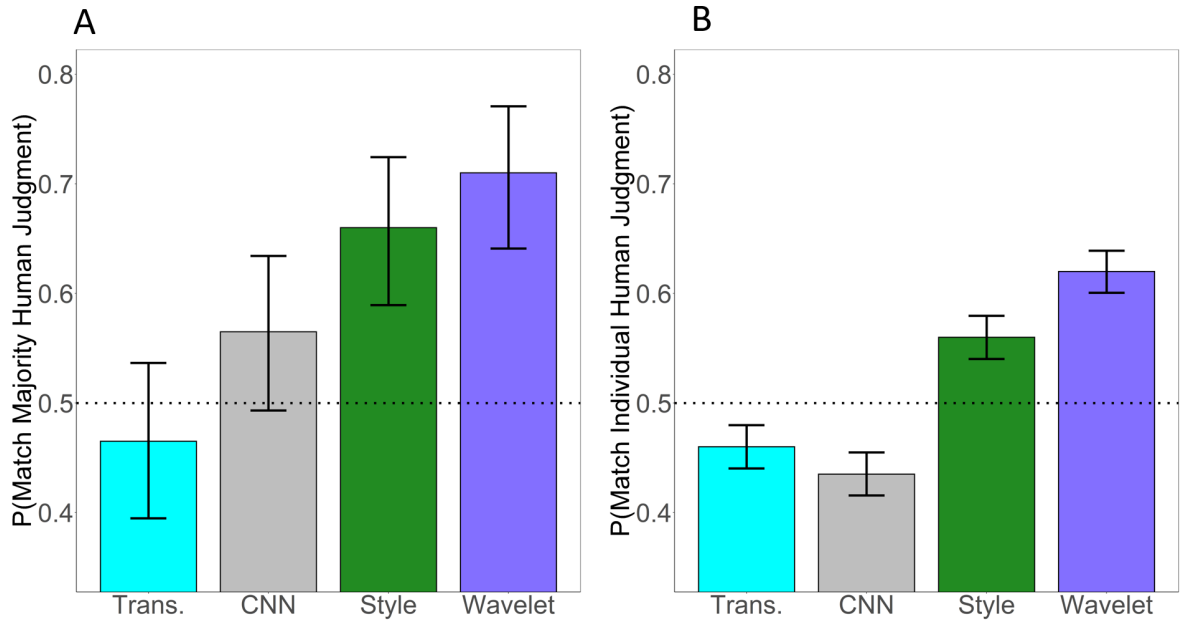

**Figure 7.** The percentage of color similarity rankings from our wavelet algorithm (purple), from convolutional DNNs trained on image classification (gray) and style transfer (green), and by a vision transformer DNN trained on image classification (cyan), that match the judgments of color similarity (A) preferred by the majority of human participants from our online survey, and (B) provided by all individual participants (i.e., not aggregated via majority ranking). There are 200 majority human judgments against which each algorithms' judgments for the same color tiles are compared, and there are 2,500 individual-level color judgments against which each algorithms' judgments are compared at the individual level. Error bars show 95% confidence intervals.

As shown in Figure 8, the difference in convolutional DNN embedding similarity is *anticorrelated* with the percentage of humans in favor of the majority pair (Spearman $\rho = -0.21$, $p < 0.04$), while the wavelet embedding differences are significantly positively correlated (Spearman $\rho = 0.38$, $p < 0.0001$). Thus, the convolutional DNN not only fails to capture human majority judgments, but also tends to embed perceptually dissimilar colors as more alike. Vision transformer DNN embedding similarity is uncorrelated with the strength of human color judgments (Spearman $\rho = 0.01$, $p < 0.0001$), and style transfer DNN embedding similarity is only weakly correlated (Spearman $\rho = 0.17$, $p < 0.05$). Embedding distance in the wavelet algorithm is significantly more predictive of the strength of human color judgments than embedding distance in the style transfer DNN ($p < 0.001$, Student T-test, Two-tailed).

In supplementary analyses, we qualitatively assess which color tile pairs the algorithms provide different color similarity embeddings compared to human judgments (Figure S21). The convolutional DNN failure modes are often surprising; for example, it judges a pair of light brown and purple tiles as more similar than a pair of light green and light yellow, whereas humans consistently judge the green/yellow pair as more similar. The style transfer DNN behaves similarly to the convolutional DNN in this regard; for example, it yields several perceptual errors that involve the same purple/brown color tile pair. Meanwhile, the vision transformer's perceptual errors often involve the outlying color tile pairs that it embeds differently, which contain colors near the boundaries of the standard RGB gamut (e.g., cyan). Color perception for all of the algorithms we test is least accurate relative to human judgments when a pair contains a color tile of very high or low luminance.

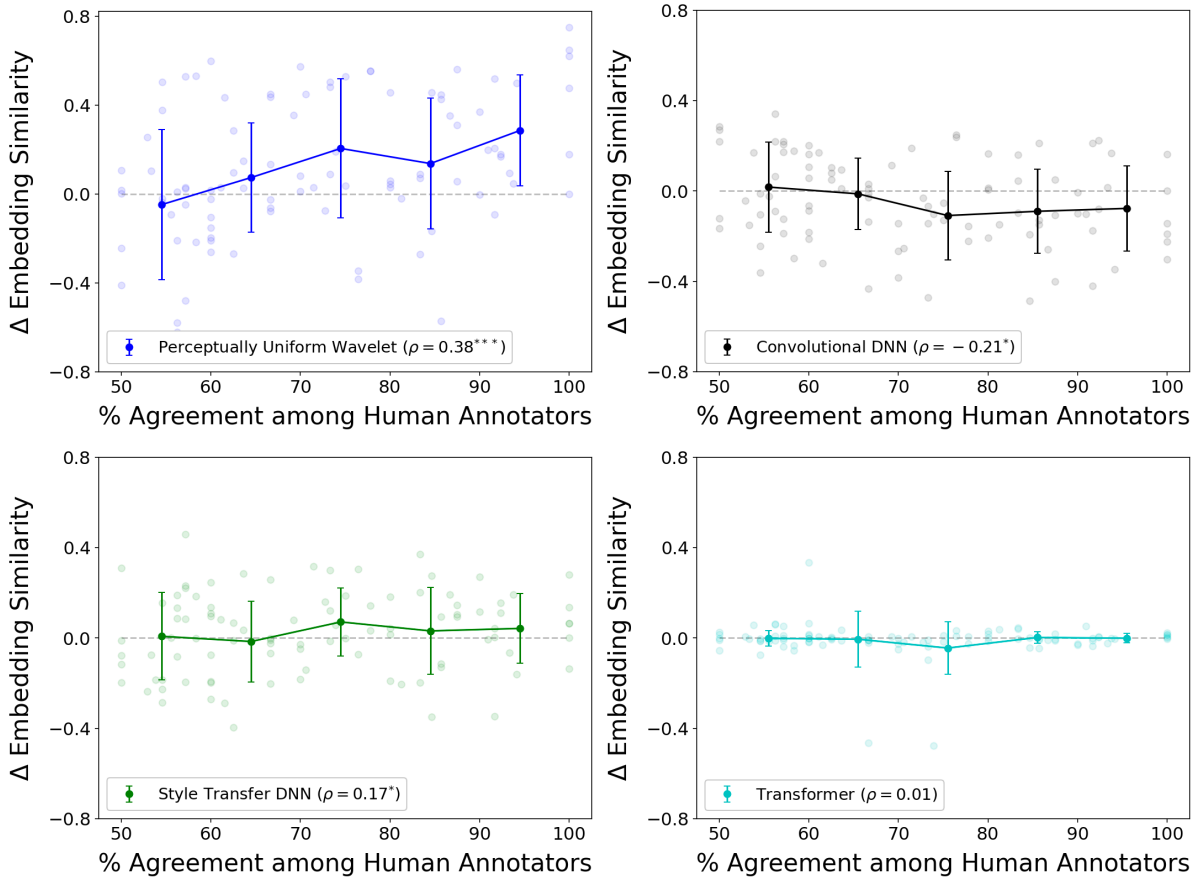

**Figure 8.** Difference between algorithmic embedding similarity for sets of two color tile pairs versus the percentage of human voters who classified the majority color pair as more similar. Wavelet, convolutional DNN, style transfer DNN, and transformer results are shown in the top-left, top-right, bottom-left, and bottom-right panels, respectively. The convolutional DNN embedding differences (black) are significantly *anticorrelated* with the strength of human color similarity agreement (Spearman ρ = -0.21, $p < 0.04$), while our wavelet algorithm's embedding differences (blue) are significantly and positively correlated with the human consensus (Spearman ρ = 0.38, $p < 0.0001$). Faded data points show individual color tile pair comparisons; bold data points show the mean, and error bars show one standard deviation.

## IV. Discussion

Although state-of-the-art deep learning algorithms often match or even surpass human-level performance on image classification tasks, we find that their perceptual representations fail to resemble or predict human color perception, which is a foundational aspect of human visual cognition. That state-of-the-art DNNs fail to accurately model human color perception casts doubt on the ability of these algorithms to serve as cognitively plausible models of human vision, particularly given the psychological importance of color perception in human vision and human cognition more generally.

Qualitatively, we find that DNNs often group images containing similar shapes (e.g., circles) or textures (e.g., checkered patterns) regardless of these images' color properties. These algorithms appear to be biased toward embedding textural and spatial information at the expense of retaining perceptually-relevant information like images' basic color properties. Intriguingly, we find that the style transfer DNN more effectively leverages color than either convolutional or vision transformer DNNs trained to classify images. Its performance is especially impressive given that this algorithm was trained on pixel-level information. The style transfer DNN's performance edge is consistent with the intuition that color is more relevant for the artistic tasks and the datasets on which style transfer DNNs are trained (e.g., the classification of paintings)[49,50,58]. This suggests that DNNs' training objectives can contribute to their ability to accurately represent color. Nonetheless, our wavelet algorithm outperforms even the style transfer DNN in predicting human color judgments (particularly at the individual level), suggesting that — even when color is more directly relevant to their training goals — the DNNs we examine have not effectively learned to retain color information. Under several analyses, we found that the ability for the style transfer DNN to cluster images based on color was indistinguishable from a wavelet algorithm with access to only grayscale information, underscoring the extent to which deep learning methods can mismodel human color perception, even when their training objectives involve color. In this sense, our results complement previous findings that extant DNN architectures may not be well-designed to retain color information throughout the visual learning process[11,27,44]. We emphasize that a systematic study of how color perception depends on network architecture and training objectives is an important area for future work.

Our work also provides insights into the effects of DNN architecture on color perception, which may reveal new pathways for improving their perceptual capabilities. Prior work suggests that DNNs drop color information in their later layers as they develop increasingly abstract representations that generalize across sensory data (e.g., learning categories of objects that often vary in their color, such as *cars* and *dogs*)[11,27,51]. Yet, this prior work primarily examines DNN color representations of real-world images; in our supplementary analysis that examines how different layers of DNNs represent controlled color stimuli (Figures S13 and S17), we find that the later layers of our convolutional DNN yield *more* perceptually coherent color representations, hinting that hierarchical learning might improve DNNs' abilities to represent basic, sensory data in certain environments. Intriguingly, a recent study shows that DNNs can preserve color information throughout their layers when trained to learn color as a categorical variable (i.e., by learning the standard English vocabulary of color names mapped to color space), showing how task structure can shape the ability for hierarchical learning to detect and preserve color information in a multimodal fashion[68]. While this approach is limited by the mapping between conventional linguistic categories and human color perception[31,69,70], which is both approximate (e.g., users of the same language can map different colors to the same color words[71]) and culturally dependent[72,73], it nevertheless highlights how DNNs can, in principle, hierarchically learn abstract associations while preserving basic perceptual information in a given sensory domain.

A promising direction for future research is to integrate feed-forward neural network architectures with components that explicitly retain sensory information and abstract representations that are characteristic of embodied cognition[35,36,38]. Progress in this direction may be aided by designing DNNs to better approximate the neurophysiology of human vision[11,18]. A rich body of work demonstrates that human vision involves separate pathways for sensory processing and the formation of abstractions from visual data (such as object categories)[74,75]; crucially, research indicates that these separate pathways remain connected throughout the human brain, enabling sensory-infused mental imagery to play an important role in domains from abstract concept learning to language comprehension[76–78]. In this context, the success of our cognitively-grounded wavelet algorithm in predicting human color judgments has several important implications. Our results provide novel empirical support for the longstanding claim in computational neuroscience that wavelet algorithms can accurately model how the brain represents color and texture[25,52,53,55]. A crucial feature of our algorithm is its transparency: the color representations formed by our wavelet algorithm are readily interpretable, in contrast to the obscurity that continues to characterize DNNs' internal representations. Thus, our wavelet algorithm provides a replicable benchmark against which color representations of future computer vision models can be compared. We posit that integrating such neurophysiologically-inspired techniques into deep-learning methods may improve their plausibility as models of human vision.

Finally, we remark that extant DNNs may not effectively leverage color because this kind of perceptual information is not necessary for popular training tasks like image classification. However, in light of numerous studies indicating that humans use color to understand both concrete and abstract concepts[31,38,39], this raises the question of how and why human cognition retains and leverages color throughout the formation of visual abstractions[4,35]. Our study is intended to serve as a springboard for future research aimed at solving this puzzle: how can deep learning architectures and training goals be modified to better retain the low-level perceptual representations that characterize human vision — and embodied cognition more broadly — without sacrificing accuracy on standard computer vision tasks such as image classification?

**Data Availability**: All code used in this study is publicly available at https://github.com/eonadler/cv-color-perception. All image clusters from our analysis, for our four main datasets (block, stripe, colorgram, and CIFAR-10 images) and four algorithms (perceptually uniform wavelet, convolutional DNN, style transfer DNN, and vision transformer DNN) are publicly available at this online repository.

**Acknowledgements**: The authors gratefully acknowledge the support of the Complex Systems Summer School hosted at the Institute of American Indian Arts and the Santa Fe Institute. We thank Aabir Kar, Sean McLaughlin, Melanie Mitchell, Ruggerio Lo Sardo, Nicholas Guilbeault, and Krishna Savani for helpful discussions and comments.

**Credit Author Statement**
E.N. and D.G. designed the project; E.N., E.D., B.D., C.C., M.C., T.H., and D.G. developed the algorithmic methods, analyzed the data, and wrote the manuscript.

Supplementary Information for:

# Divergences in Color Perception between Deep Neural Networks and Humans

This appendix contains:
    Supplementary Materials and Methods
    Supplementary Analyses
    Supplementary References

**Supplementary Materials and Methods**

Here, we provide additional details on the image datasets, computer vision algorithms, and methods for analyzing image embeddings used in our main study.

**Image Datasets.**

- *Block and Stripe Images*: We use the Python Imaging Library PIL to generate block and stripe images. In particular, we create two sets of 1000 (300 x 300)-pixel images composed of either one randomly-selected color (the "block" dataset), or two randomly-selected colors arranged in a vertical stripe pattern (the "stripe" dataset). In both cases, a white border is included around a square of the primary color(s); this border is not used when calculating color similarity. In supplementary analyses, we show that different background colors do not significantly affect block image embeddings (Figure S20).

- *Colorgrams*: To generate colorgrams, we use composite Google Image search results developed using the comp-syn package[1,2]. Specifically, colorgrams are generated by averaging the top 100 Google Image results for a given search term. Because Google Image results are driven by search and usage popularity according to the PageRank algorithm[3], each colorgram can be interpreted as an average visual representation of its corresponding concept according to Google Images. We select 1,000 colorgrams from the dataset of 40,000 colorgrams in Desikan et al. (2020)[1], corresponding to the most frequently used words in English.

- *CIFAR-10*: We collect 10,000 images from the standard CIFAR-10 dataset; we also collect higher-resolution versions of these images that have been resized using the CAI Super Resolution technique.

**Computer Vision Algorithms.**

- *Perceptually Uniform Wavelet Algorithm*: We implement our wavelet algorithm using the wavelet scattering transformation package Kymatio (Andreux et al. 2018)[4]. Our wavelet transforms consist of convolutions with the wavelet filter followed by complex modulus operations. Specifically, $J = 5$ wavelet scales and $L = 4$ orientations are computed at each level. The resulting zeroth-order coefficient corresponds to a single convolution with a low-pass filter; the first-order coefficient corresponds to a convolution with a wavelet $\Psi_j^l$ with scale $j = 1, \ldots, J$ and orientation $l = 1, \ldots, L$ followed by a complex modulus operation and convolution with the low-pass filter; and the second-order coefficient is a convolution of two wavelet filters with $j_2 > j_1$ interleaved with the complex modulus operation and followed by a convolution with the low-pass filter. This yields a total of $1 + JL + [L^2(J-1)J]/2$ coefficients, which we reduce to $1 + J + [J(J-1)]/2 = 16$ dimensions by averaging over the $L$ orientations.

  When operating our wavelet algorithm in $J_zA_zB_z$ color space, we calculate $J_zA_zB_z$ pixel values using the procedure described in Safdar et al. (2017)[5], and we run the wavelet transform on each color channel individually, yielding 3 x 16 = 48 coefficients that we combine into a single embedding. In certain tests, we run our algorithm on grayscaled versions of each image with 16 coefficients each to study the impact of image hue and saturation.

- *Convolutional DNN*: We use the ResNet model as implemented in the img2vec Python package to create convolutional DNN embeddings for each image dataset. In **Study 1: B. Clustering in Perceptually Uniform Color Space**, we describe our procedure for extracting embeddings from different ResNet layers as a robustness test.

- *Style Transfer DNN*: We use the style transfer algorithm described in Ghiasi et al. (2017)[6], which is implemented in TensorFlow[7] and uses the VGG19 convolutional DNN architecture. In particular, we extract 4096-dimensional image embeddings by flattening the (64 x 64) block matrix from the penultimate layer of this network.

- *Vision Transformer DNN*: We use the final embedding layer of the algorithm from Dosovitskiy et al. (2021)[8]; in particular, this layer uses information from the 196 locational "patch" embeddings, which were originally optimized to produce Imagenet[9] label classifications.

**Analysis Techniques.**

- *Clustering Techniques*: We use the [sklearn.cluster.KMeans](sklearn.cluster.KMeans) package with init='k-means++', which performs several random trials at each clustering step to ensure convergence in the final clustering assignments.

- *Color Coherence Measure*: We measure the color coherence of the image groups returned by *k*-means clustering as follows. Let $c(x,y)$ denote an image, where $(x,y)$ are image coordinates and $c(x,y)$ is the three-dimensional color tuple of each pixel in the image (we do not operate on the alpha channel for any of the image sets). We transform these tuples to perceptually uniform $J_z A_z B_z$ color space[5] and measure $J_z A_z B_z$ distributions, concatenated over pixel coordinates, using eight evenly-segmented $J_z A_z B_z$ subvolumes following. We quantify the similarity between the color distributions of pairs of images in each cluster as follows. Let $c_i(x,y)$, $c_j(x,y)$ denote images *i* and *j*, with $J_z A_z B_z$ distributions $c_i$ and $c_j$, respectively. In particular, we define the color similarity

$$C(c_i \,||\, c_j) = 1 - D_{JS}(c_i \,||\, c_j) = 1 - [D_{KL}(c_i \,||\, c_{ij}) + D_{KL}(c_j \,||\, c_{ij})]/2, \quad (1)$$

where $D_{JS}$ ($D_{KL}$) denotes Jensen-Shannon (Kullback-Leibler) divergence and $c_{ij} = (c_i + c_j)/2$. Note that larger values of $C(c_i \,||\, c_j)$ correspond to image pairs with more similar color distributions.

# Supplementary Analyses

## Study 1: A. Visual Properties of Image Clusters

*Additional Image Cluster Examples*

All image clusters for our main datasets (i.e., stripe images, colorgrams, and CIFAR-10 images) and algorithms (i.e., our perceptually uniform wavelet and convolutional, style transfer, and vision transformer DNNs) are provided at the following link: image cluster repository. This dataset and all supporting data and code will be made publicly available upon publication.

Visual inspection of these image clusters supports our main quantitative results. In particular, 1) the images grouped by our perceptually uniform wavelet algorithm tend to have noticeably more similar color properties, 2) the images grouped by our convolutional and vision transformer DNNs tend to exhibit similar textural patterns at the expense of discriminating between perceptually similar colors , and 3) the images grouped by our style transfer DNN tend to compromise between color and textural similarity relative to the other DNNs. Examples of clusters returned by each algorithm for our colorgram analysis are shown in Figures S1-S4. Interestingly, we note that the style transfer clusters often capture conceptual similarities among images, e.g., by grouping colorgrams representing Google Image searches of food terms.

Similarly, visual inspection of the stripe images returned by our clustering analysis confirms many of the trends identified in our analyses. Specifically, Figures S5-S8 show that stripe images grouped by our perceptually uniform wavelet algorithm tend to have similar hues and luminosities, while the convolutional and vision transformer DNNs can group images of surprisingly different colors (e.g., the convolutional DNN groups red-green and nearly black stripe images with bright purple-ish hues). Meanwhile, the style transfer DNN often groups images with colors of similar luminance that are not perceptually similar when taking hue and saturation into account.

*Properties of Wavelet Analysis Clusters*

To further characterize the properties of image clusters returned by our perceptually uniform wavelet algorithm, we perform principal component analysis on our wavelet algorithm's coefficients, and identify the images that minimize and maximize principal components that explain the majority of the resulting variance. The results of this analysis are shown in Figure S9 and indicate that our wavelet algorithm's coefficients *simultaneously* capture images' color and textural properties. For example, images that minimize and maximize wavelet coefficients in the bottom rows of Figure S9 are differentiated by both hue (e.g., whitish/grayish/blackish vs. blueish/yellowish hues) and shape (e.g., square vs. circular patterns). A detailed study of the image properties captured by our wavelet coefficients, and the way in which these coefficients change when using different color space representations, is therefore an interesting avenue for future work.

**Perceptually uniform wavelet colorgram cluster**

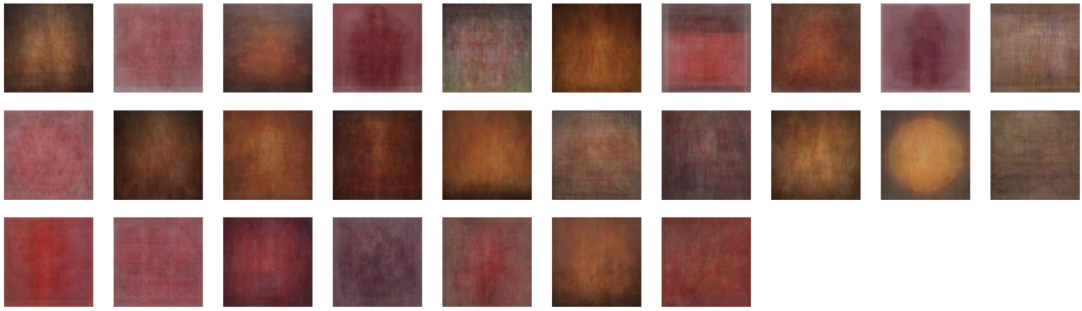

**Figure S1.** Example of a colorgram cluster returned by our wavelet algorithm. The images have similar color distributions, even in perceptually uniform color space.

**Convolutional DNN colorgram cluster**

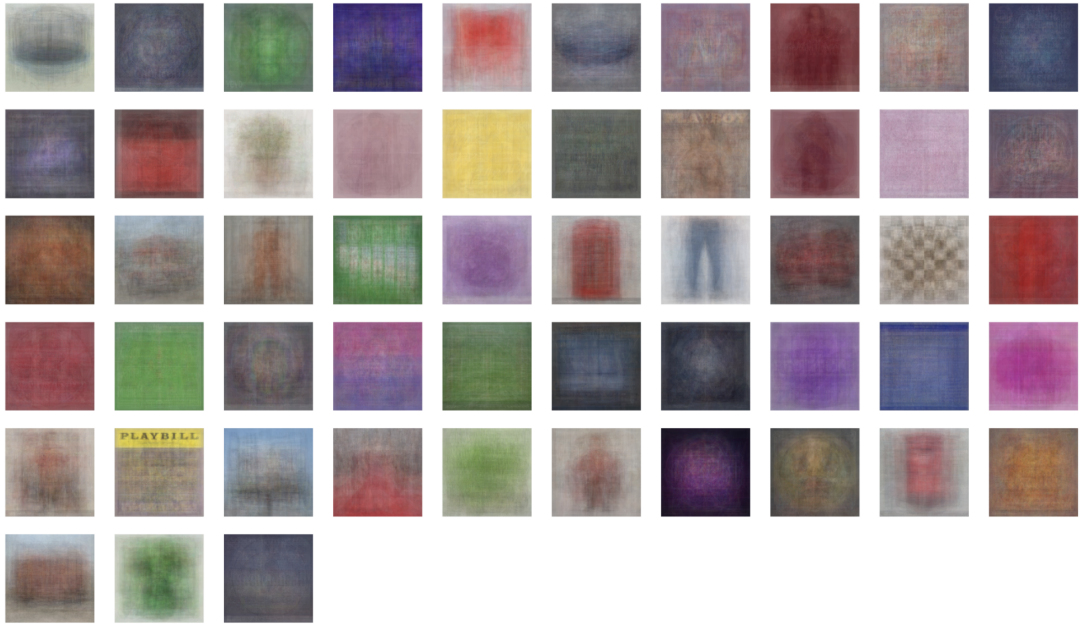

**Figure S2.** Example of a colorgram cluster returned by our convolutional DNN. The images have widely varying color distributions, though some exhibit similar textural patterns (e.g., squares and circles).

**Style transfer DNN colorgram cluster**

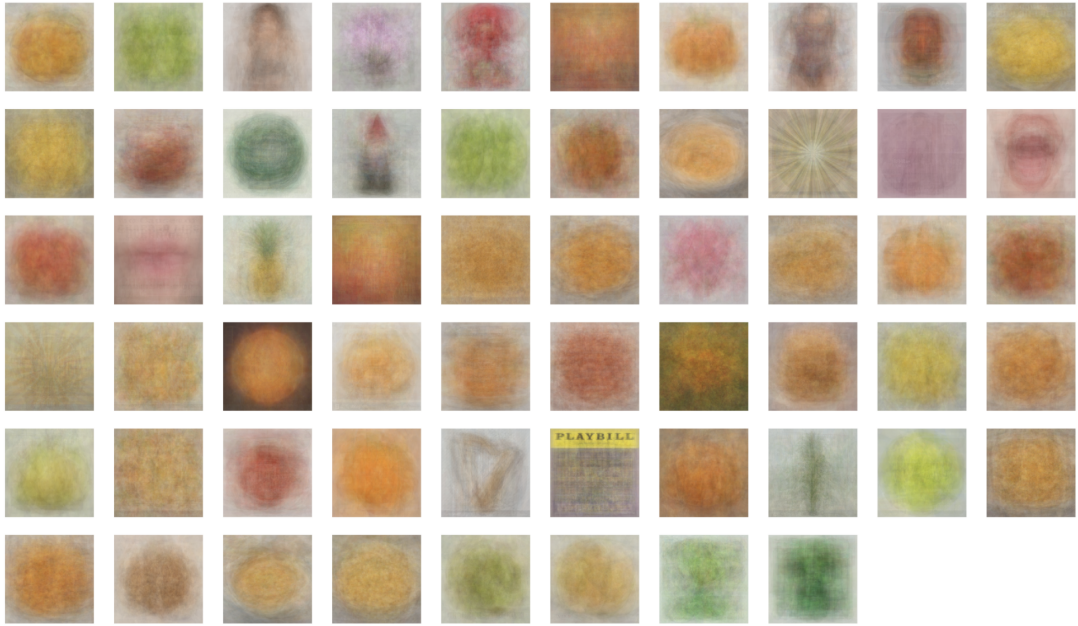

**Figure S3.** Example of a colorgram cluster returned by our style transfer DNN algorithm. This cluster contains several colorgrams resulting from Google Image searches of food terms (e.g., "pineapple").

**Vision transformer DNN colorgram cluster**

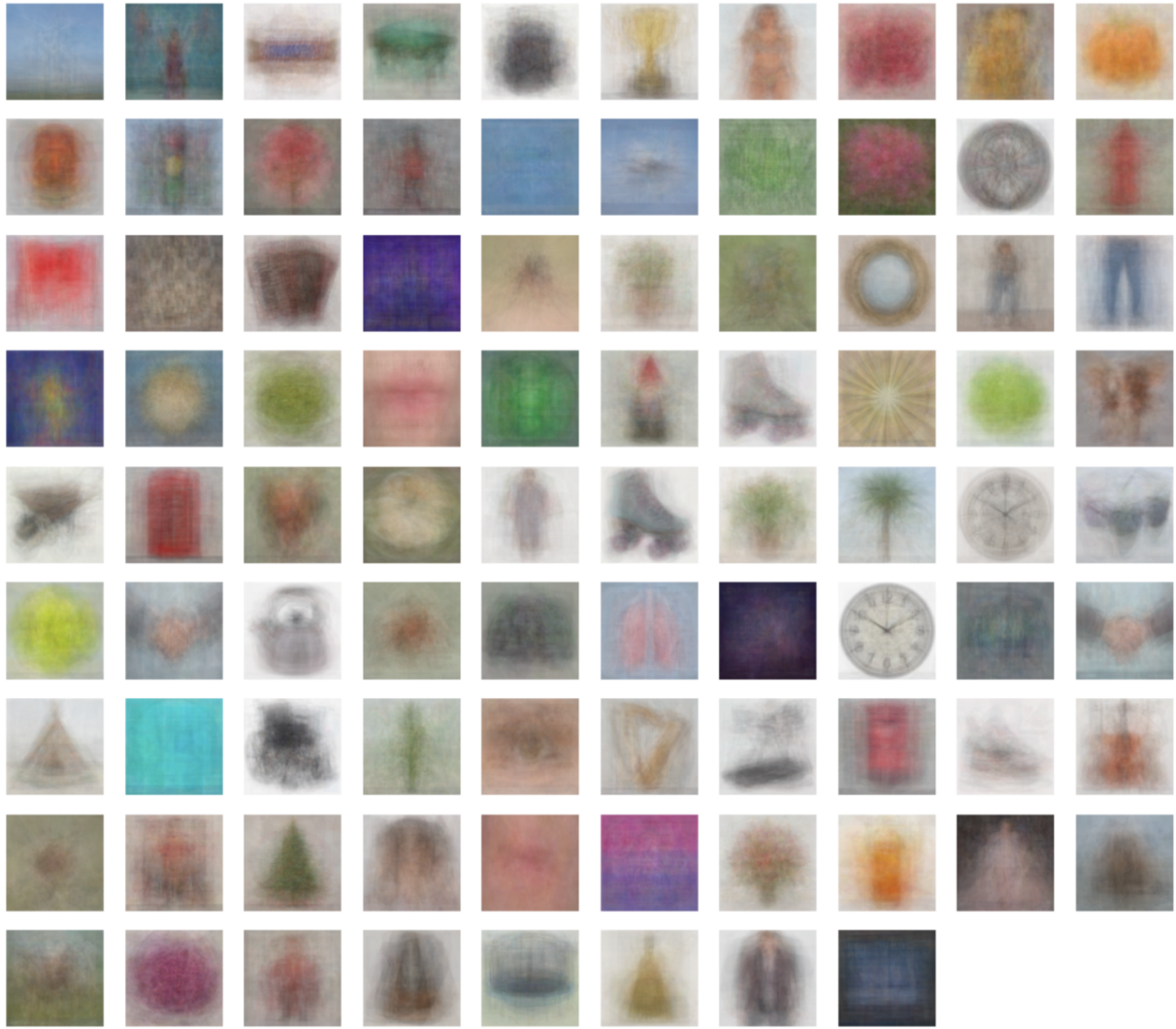

**Figure S4.** Example of a colorgram cluster returned by our vision transformer DNN algorithm.

**Perceptually uniform wavelet stripe cluster**

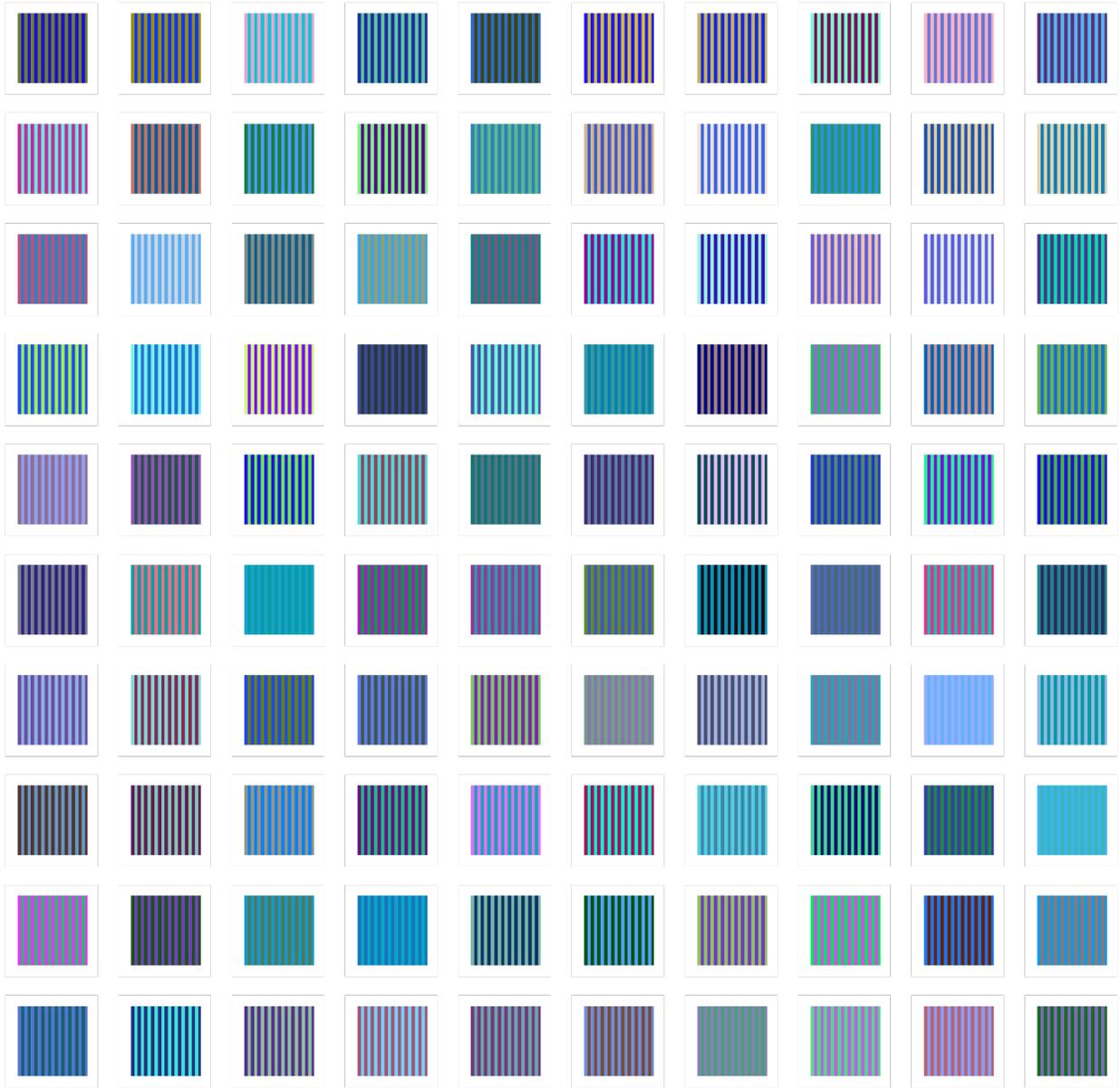

**Figure S5.** Example of a stripe image cluster returned by our perceptually uniform wavelet algorithm. The images have similar color distributions, even in perceptually uniform color space.

**Convolutional DNN stripe cluster**

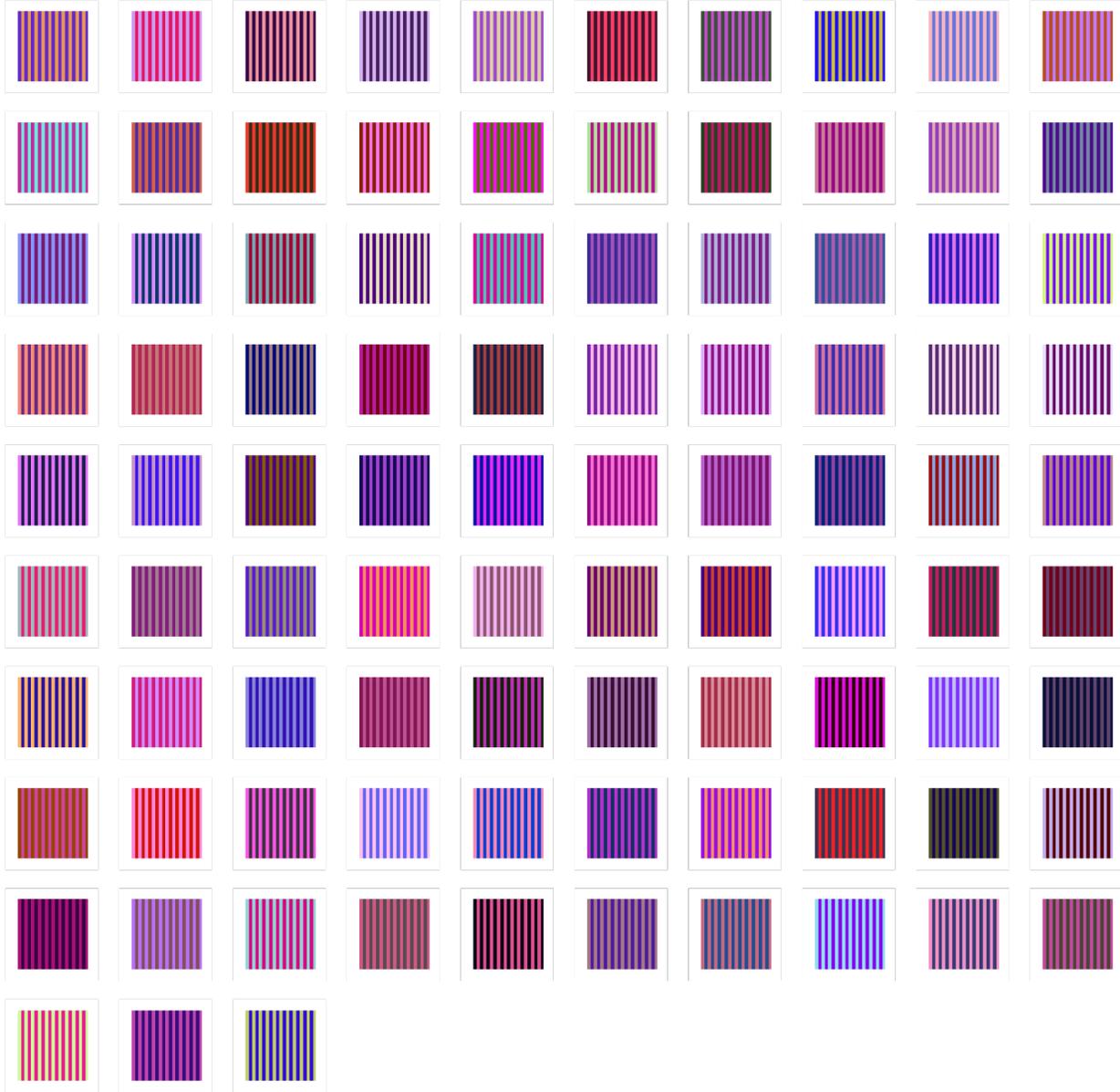

**Figure S6.** Example of a stripe image cluster returned by our convolutional DNN algorithm. Several images have widely varying color distributions; these perceptual errors often involve images with high or low luminance (e.g., the black-ish image near the bottom right falls in the same cluster as bright purple images).

**Style transfer DNN stripe cluster**

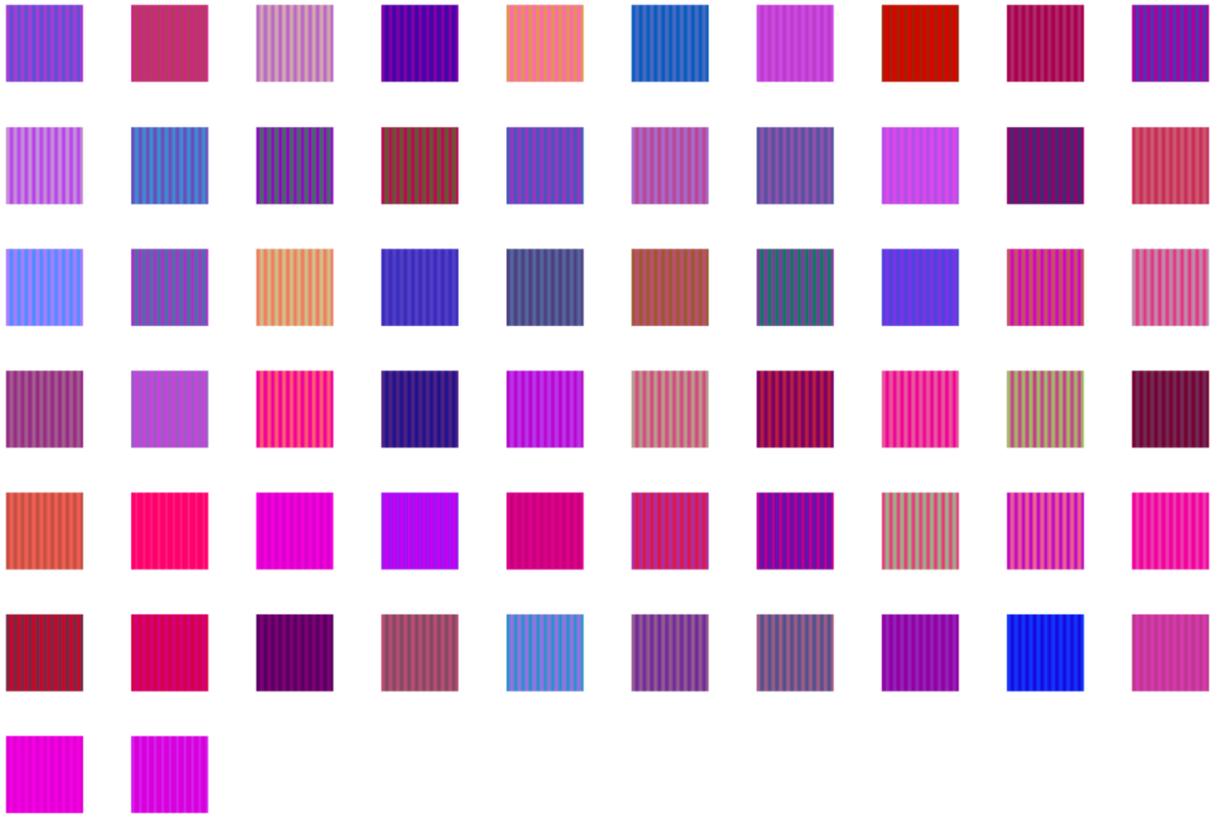

**Figure S7.** Example of a stripe image cluster returned by our style transfer DNN algorithm.

**Vision transformer DNN stripe cluster**

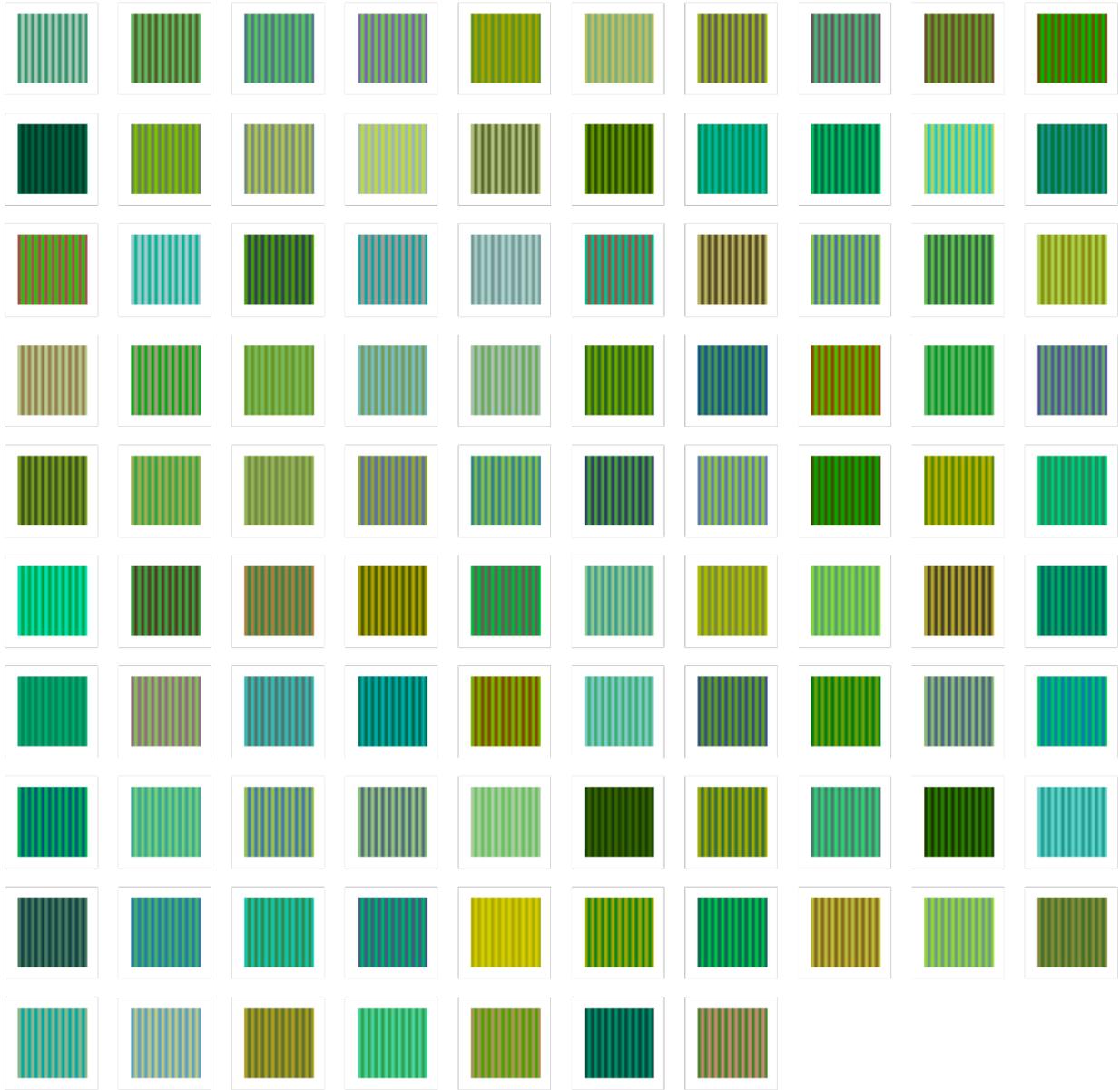

**Figure S8.** Example of a stripe image cluster returned by our vision transformer DNN algorithm.

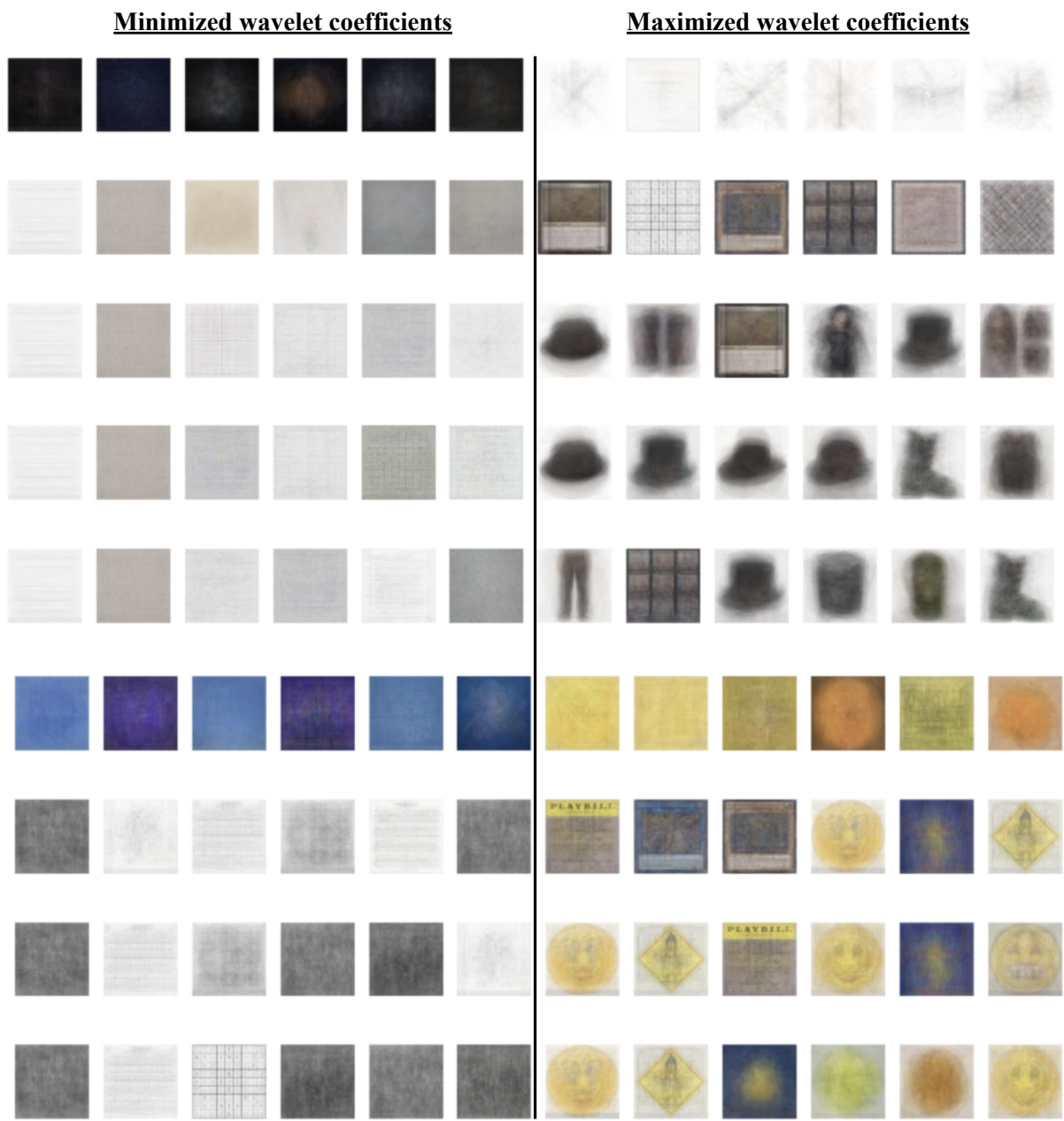

**Figure S9.** Examples of colorgrams that minimize and maximize principal components of our perceptually uniform wavelet algorithm's embedding coefficients. Each row corresponds to a different wavelet coefficient that is either minimized (to the left of the vertical line) or maximized (to the right of the vertical line). Wavelet coefficients simultaneously differentiate images by luminosities, textural properties (e.g., checkered patterns), and hues, indicating that our algorithm captures both color and textural properties of images.

## Study 1: B. Clustering in Perceptually Uniform Color Space

*Additional Color Clustering Results*

Figures S10-S11 illustrate the color space distribution of image clusters returned by our perceptually uniform wavelet and convolutional, style transfer, and vision transformer DNN algorithms for our colorgram and CIFAR-10 datasets, in the same format as Figure 3, which showed results for our stripe image dataset. The qualitative results shown in Figure 3 all hold for the other datasets; in particular, image clusters returned by our perceptually uniform wavelet algorithm are more clearly separated in color space than those returned by the convolutional, style transfer, or vision transformer DNNs. This separation is more evident for colorgrams than for CIFAR-10 images, which is consistent with the color clustering results reported in the main text.

The mean colors of the image clusters shown in Figures S10-S11 are also consistent with our finding that stripe images, colorgrams, and CIFAR-10 images successively display narrower ranges of image hues and luminosities. The average colors of CIFAR-10 images are particularly muted, most often sampling grayish, brownish and dark blueish hues, which is likely due to the color averaging effects of "background" pixels in real-world images. For CIFAR-10 and other image classification datasets, these foreground objects are members of the image classes of interest. Meanwhile, "background" pixels are known to influence DNN image classifiers[10]. Thus, the color properties of these "background" pixels may also affect image embeddings and artificially reduce the importance of color for accurately representing and differentiating foreground objects.

*Comparison between $J_zA_zB_z$ and RGB Wavelet Algorithms*

Figure S12 compares image clusters from our fiducial wavelet algorithm, which uses the approximately perceptually uniform $J_zA_zB_z$ color space, to the results of the same algorithm run on images represented in RGB color space. We show this comparison for our stripe and colorgram datasets, since these algorithms' CIFAR-10 clustering results do not significantly different (Figure 4B). The visual properties of these algorithms' image clusters are similar, supporting our finding that the $J_zA_zB_z$ wavelet's success is not solely driven by the fact that it represents images in an approximately perceptually uniform color space.

*Robustness to DNN Embedding Layer*

To assess whether the DNN embedding layers we use in our analyses affect our color coherence results, we rerun our analyses using the first, third ("conv3_block4_out"), and fifth ("conv5_block4_out") layers of our ResNet network ("convolutional DNN"), and compare these to our fiducial results that use its penultimate ("avg_pool") layer. The convolutional DNN is well suited for this test because, unlike our style transfer DNN, its embeddings are not themselves aggregates of multiple embedding layers.

The clustering result for our stripe image dataset is shown in Figure S13. Visually, color clustering fidelity improves toward deeper embedding layers; we confirm this by calculating the color coherence fraction, which we find to be $f_{first} = 0$, $f_{third} = 0.12$, and $f_{fifth} = 0.21$, compared to $f_{CNN} = 0.22$ using the penultimate layer from our main analysis. Note that the first layer embeds all stripe images similarly, such that our *k*-means algorithm returns only one distinct cluster; we therefore assign this result a coherence fraction of zero. Meanwhile, the fifth layer is nearly indistinguishable from the penultimate layer in terms of both the coherence fraction and the visual properties of the clusters in Figure S13.

Thus, earlier layers of the convolutional DNN perform more poorly than the penultimate layer we use throughout, which in turn performs much more poorly than our wavelet algorithm, which returns $f_{wavelet}$ = 0.81 on the same stripe dataset. In SI Section **C. Color Coherence of Image Clusters**, we demonstrate that the average values of the resulting color similarity distributions are statistically consistent for all layers we examine, with the caveat that the first layer only returns one cluster. Thus, our main conclusions are robust to the specific embedding layers we extract.

*Robustness to Training Objective*

To assess whether DNNs' training objectives affect the fidelity of their color perception, we test a convolutional DNN based on the U-Net architecture that is trained to perform semantic segmentation rather than classification[11]. Note that the encoder portion of the U-Net architecture is similar to the ResNet model ("Convolutional DNN") we study throughout the paper. To compare with our fiducial Convolutional DNN results, which use the penultimate embedding layer, we extract the penultimate embedding layer from the U-Net trained for image segmentation. This layer corresponds to a 224 x 224 matrix, which represents transformed images after convolutions are applied. We unravel this matrix into a 50,176-dimensional vector that we use to reproduce our clustering analysis.

The clustering result for our stripe image dataset is shown in Figure S14. Visually, the image segmentation DNN clusters stripe images with less fidelity than our fiducial classification DNN; we confirm this by calculating the color coherence fraction, which we find to be $f_{segmentation}$ = 0.15, compared to $f_{CNN}$ = 0.22 in our main analysis. Thus, although the image segmentation DNN performs slightly worse than the classification DNN at this task, both models' color clustering ability is well below that of our wavelet algorithm, which returns $f_{wavelet}$ = 0.81 on the same dataset. Furthermore, as we demonstrate in SI Section **C. Color Coherence of Image Clusters**, the average values of these DNNs' color similarity distributions are statistically consistent.

We therefore conclude that our main findings are robust to the specific training objectives of the networks we examine. As discussed in the main text, a style transfer training objective improves the fidelity of DNN color perception in certain regards, but even the style transfer DNN that we examine consistently and significantly underperforms compared to our wavelet algorithm. A dedicated study of the interplay between color perception, training objectives, and network architectures will be a fruitful area for future work; for example, the image segmentation network is most sensitive to the difference between warm and cool hues, which may follow because these hues often delineate images' foregrounds and backgrounds.

*Robustness to k-means Clustering Algorithm*

We also explicitly verify that our k-means clustering results do not depend on the specific algorithm we use for our main analyses. In particular, Figure S15 compares our fiducial result for style transfer DNN clustering of the stripe dataset with an alternative k-mean clustering algorithm that uses the 'random' initialization strategy with 10 initial random clusters (n_init = 10). The clustering results differ slightly at a fine-grained level, but the overall structure and all color clustering statistics are unaffected; this continues to hold for all other algorithms and datasets in our study.

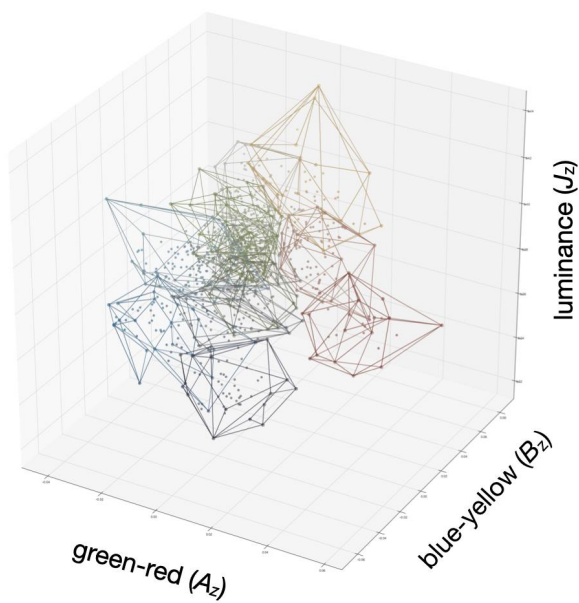
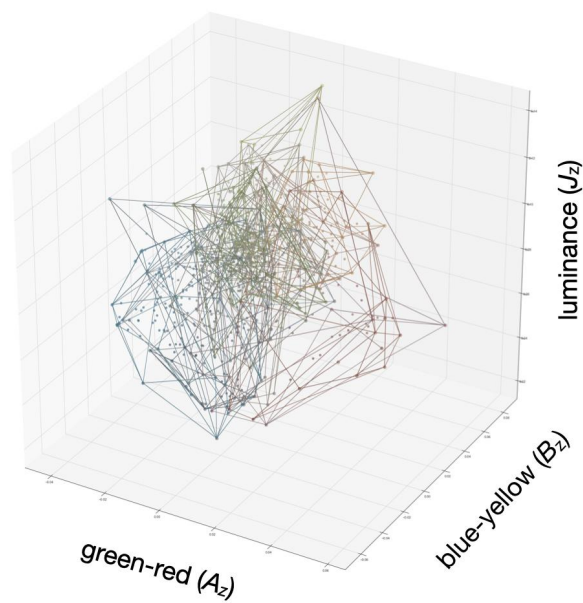
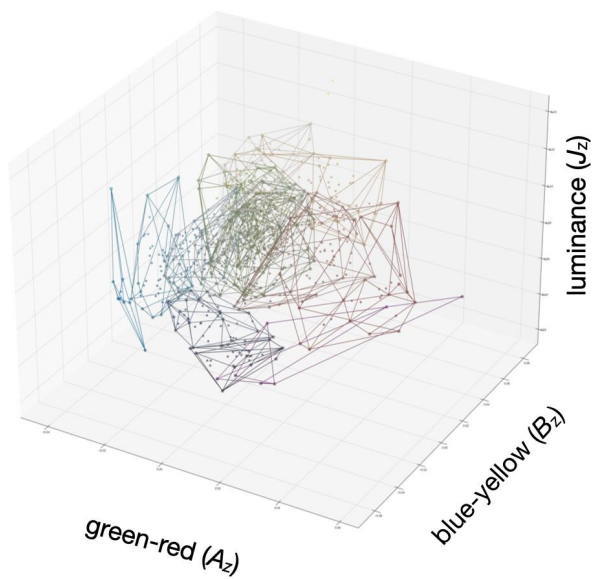
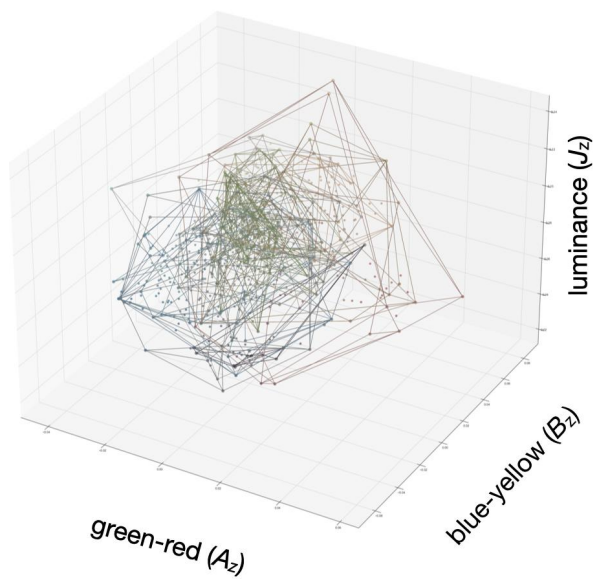

**Figure S10.** Comparison of the color properties of images clustered by our perceptually uniform wavelet (top-left) convolutional DNN (top-right), style transfer DNN (bottom-left), and vision transformer DNN (bottom-right) algorithms for images in our colorgram dataset.

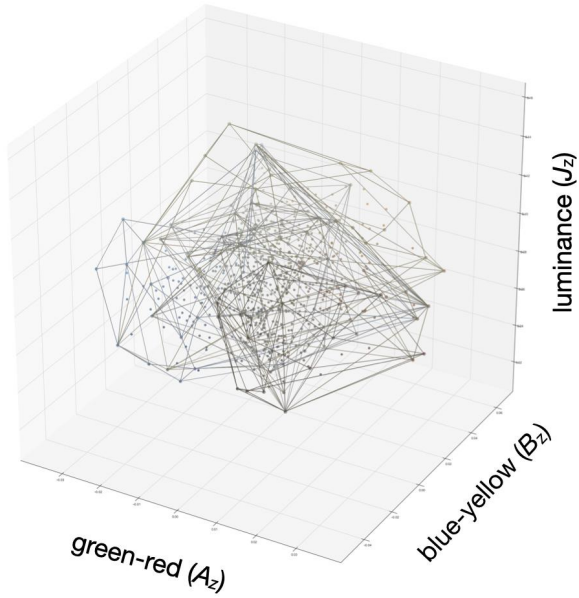 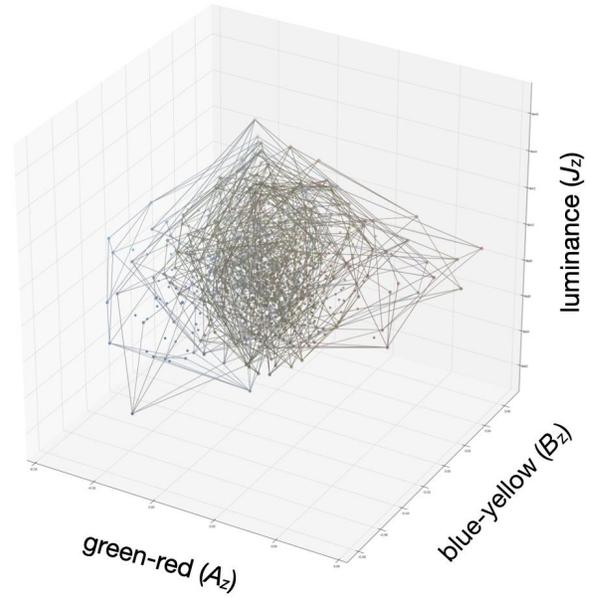
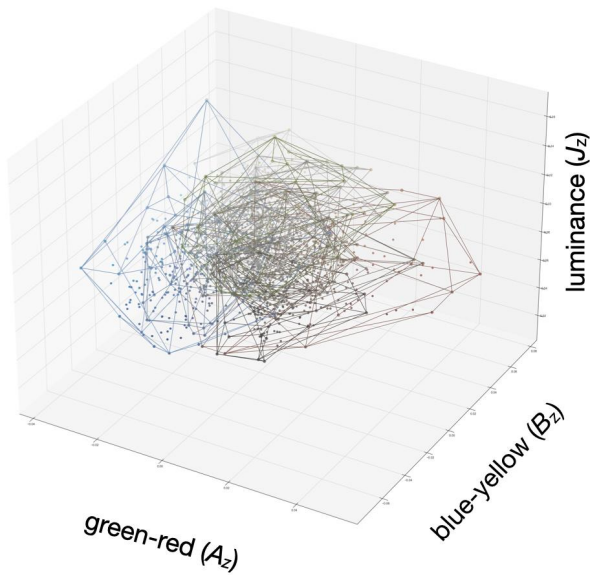 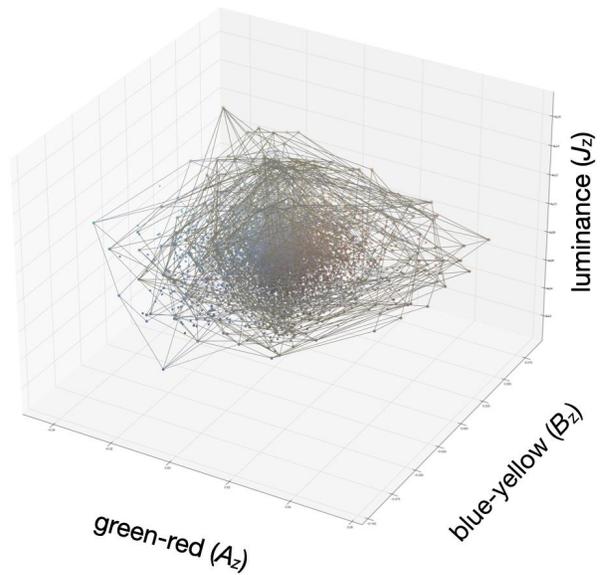

**Figure S11.** Comparison of the color properties of images clustered by our perceptually uniform wavelet (top-left) convolutional DNN (top-right), style transfer DNN (bottom-left), and vision transformer DNN (bottom-right) algorithms for images in our CIFAR-10 dataset.

## Perceptually Uniform Wavelet

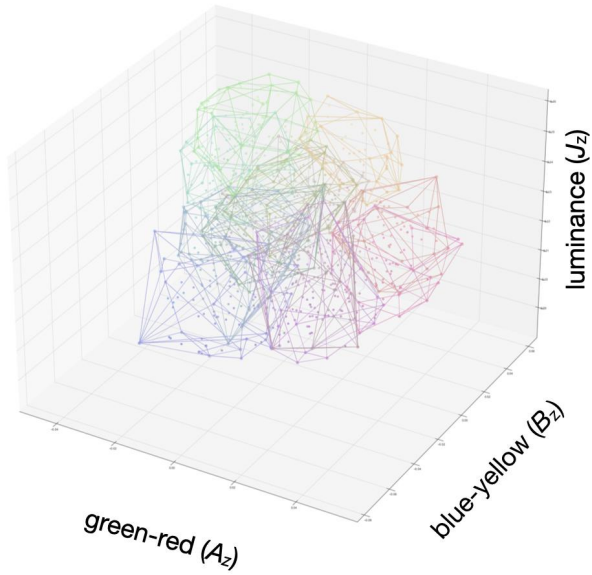

## RGB Wavelet

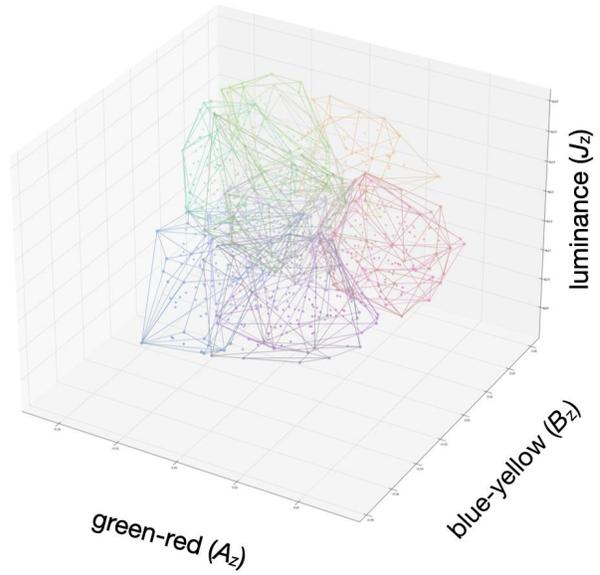

## Perceptually Uniform Wavelet

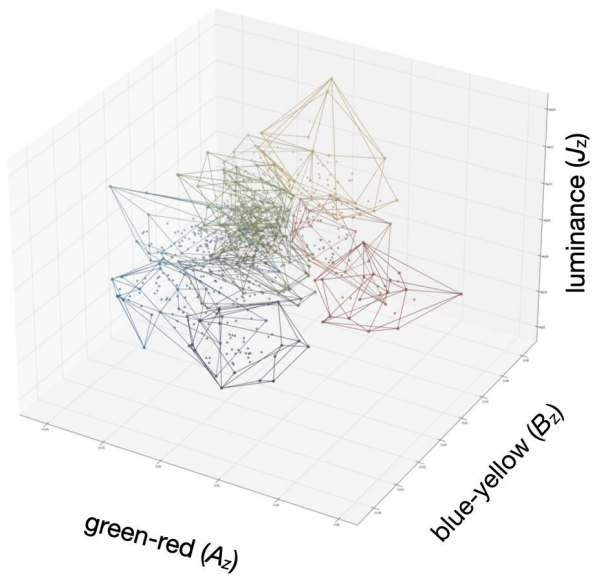

## RGB Wavelet

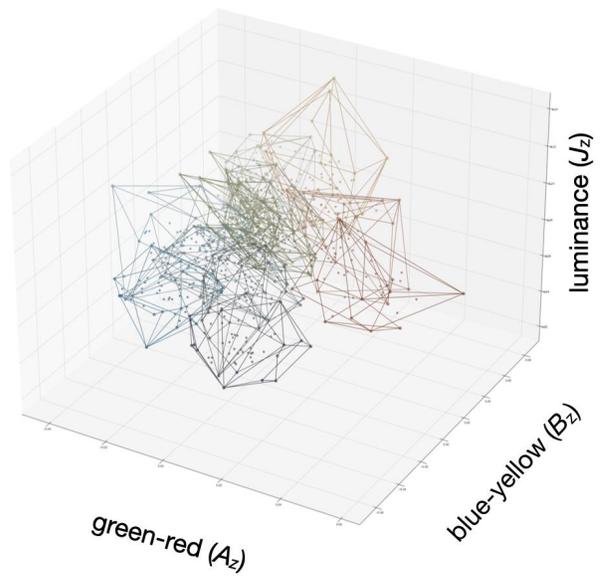

**Figure S12.** Comparison of the color properties of images clustered by our fiducial wavelet algorithm that operates in an approximately perceptually uniform color space (left column) versus the same algorithm that operates on images represented in RGB color space (right column) for our stripe (top row) and colorgram (bottom row) image datasets.

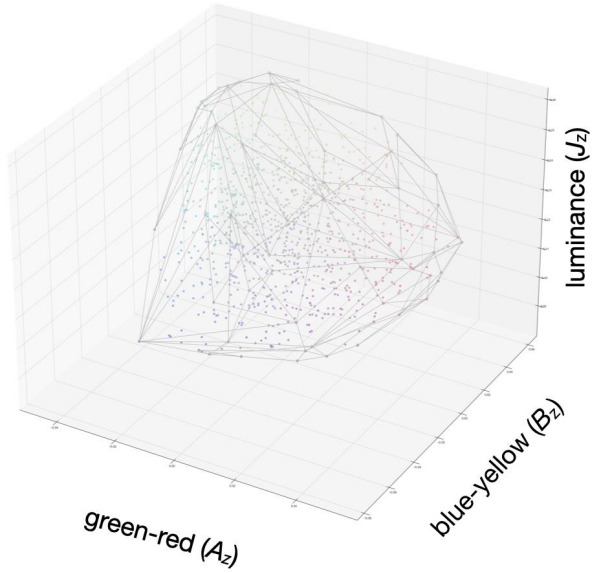
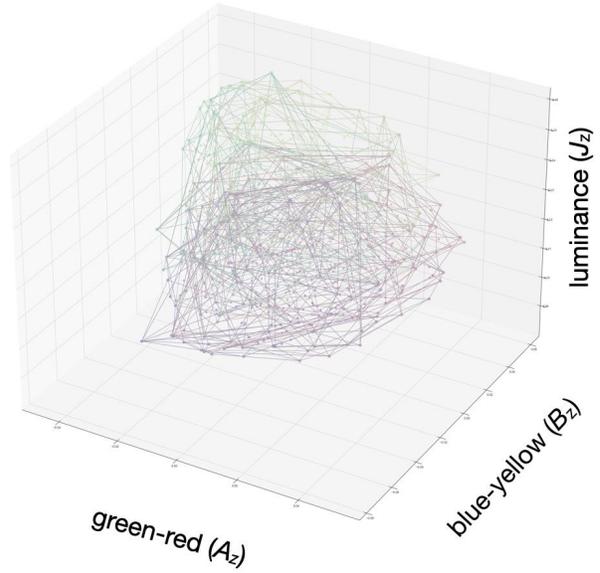
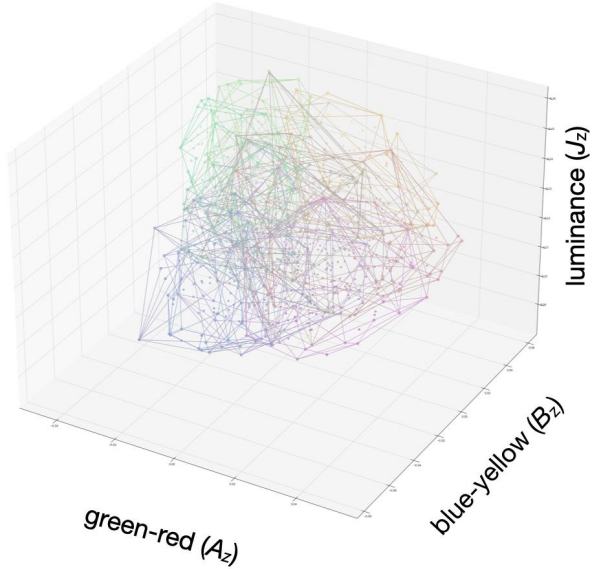
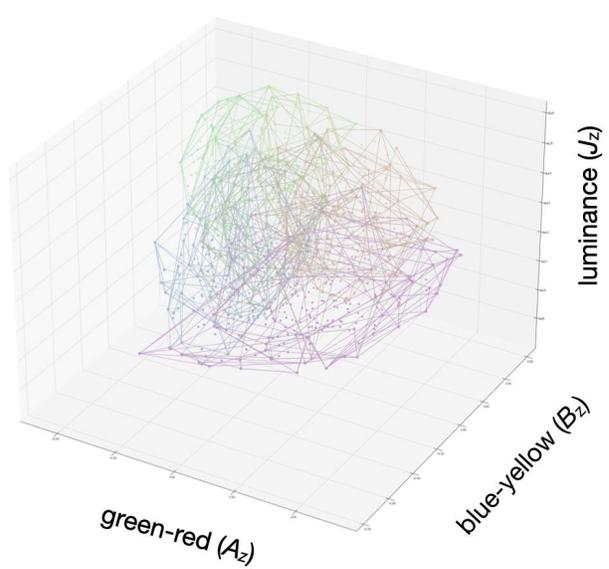

**Figure S13.** Comparison of the color properties of images clustered by our convolutional DNN algorithm using the first (top left), third (top right), fifth (bottom left), and penultimate (bottom right, our fiducial result) embedding layer. Note that the first layer only returns one cluster.

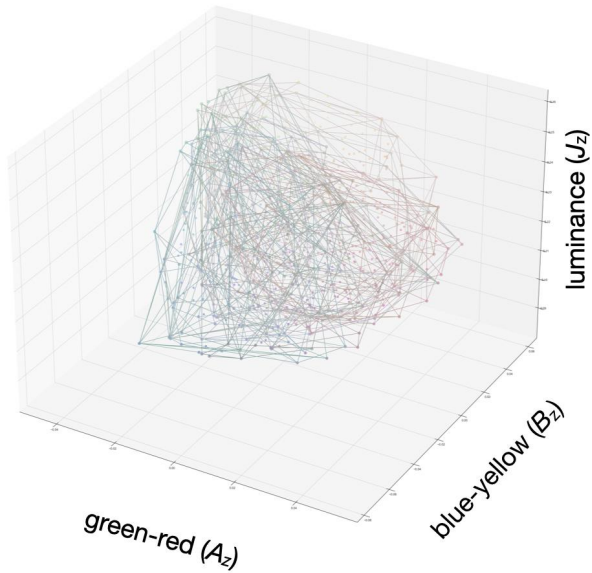
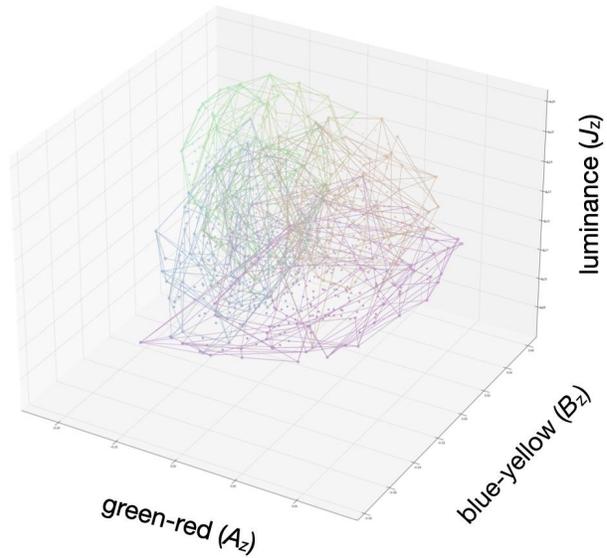

**Figure S14.** Comparison of the color properties of images clustered by a convolutional DNN trained on image segmentation using the U-Net architecture (left), versus a convolutional DNN trained on image classification using the ResNet architecture (right, our fiducial result).

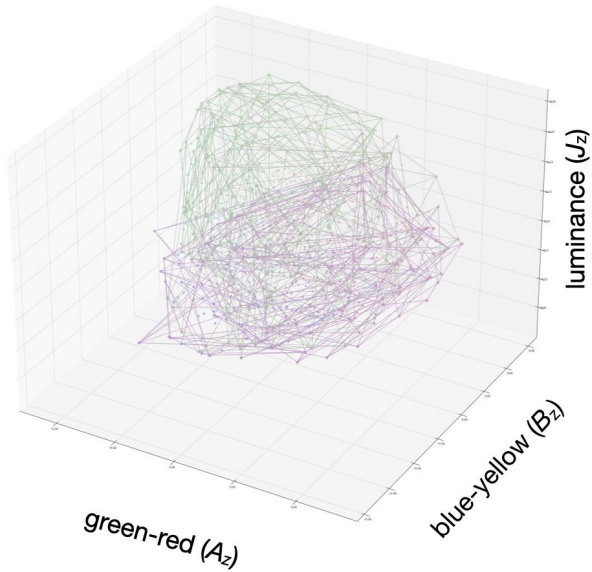
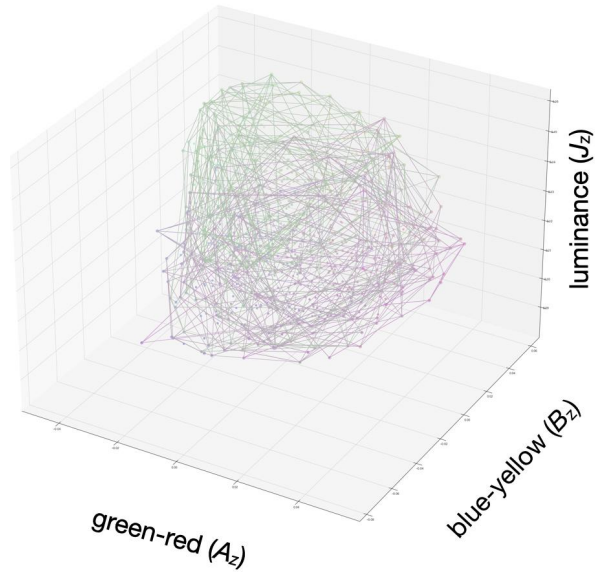

**Figure S15.** Comparison of the color properties of images clustered by our style transfer DNN algorithm on the stripe dataset, for our fiducial k-means clustering algorithm (left) versus an alternate version that uses the 'random' initialization strategy with 10 initial random clusters.

**Study 1: C. Color Coherence of Image Clusters**

*Additional Color Coherence Results*

Figure S16 shows the color similarity distributions for our stripe and CIFAR-10 datasets returned by our perceptually uniform wavelet, grayscaled wavelet, convolutional DNN, and style transfer DNN algorithms. These distributions are qualitatively consistent with our findings in the main text: the perceptually uniform wavelet algorithm yields clusters with the highest color similarity, while the DNNs yield less color-coherent clusters. Note that the stripe dataset's color similarity distribution is discontinuous because these images are composed of color pairs selected from discrete points in RGB color space. Furthermore, the variability from random clustering assignments for this dataset is large compared to the spread of the mean color coherence values among the algorithms we consider.

Both panels of Figure S16 also include the result of our wavelet algorithm when operating in RGB color space. Consistent with our results for the colorgram dataset in the main text, the wavelet algorithm's color coherence does not significantly degrade when using RGB. Note that, for stripe images, the improvement of the RGB wavelet's mean color coherence is not statistically significant given the typical spread of random clustering assignments.

*Robustness to DNN Embedding Layer and High-Level Visual Training Objective*

Following our robustness tests in **B. Clustering in Perceptually Uniform Color Space**, we verify that our color similarity results are robust to the choices of DNN training objective and embedding layer.

The left panel of Figure S17 shows color similarity distributions for our stripe image dataset returned by the first (magenta), third (orange), fifth (green), and penultimate (black) embedding layer of the convolutional DNN. The average values of these color similarity distributions are all consistent within the statistical uncertainties of clustering assignments for our stripe dataset. Furthermore, consistent with Figure S13, earlier embedding layers return lower color similarity, on average, than our result using the penultimate embedding layer. Thus, we conclude that the embedding layer we extract for our main tests does not affect our conclusions; moreover, we may slightly overestimate convolutional DNNs' color perception fidelity by using the penultimate embedding layer.

The right panel of Figure S17 shows the color similarity distribution for our stripe image dataset returned by our perceptually uniform wavelet algorithm (blue), our standard convolutional DNN trained for image classification using the ResNet architecture (black), and a convolutional DNN trained for image segmentation using the U-Net architecture (red). The average values of these DNNs' color similarity distributions are consistent within the statistical uncertainties of clustering assignments for our stripe dataset. Consistent with Figure S14, the DNN trained on image segmentation returns slightly lower color similarity, on average, than our fiducial convolutional DNN. Thus, we conclude that variations in high-level visual training objectives do not significantly affect our color similarity distributions. We have made the code for our embedding layer and training objective tests publicly available, along with the rest of our materials, at https://github.com/eonadler/cv-color-perception.

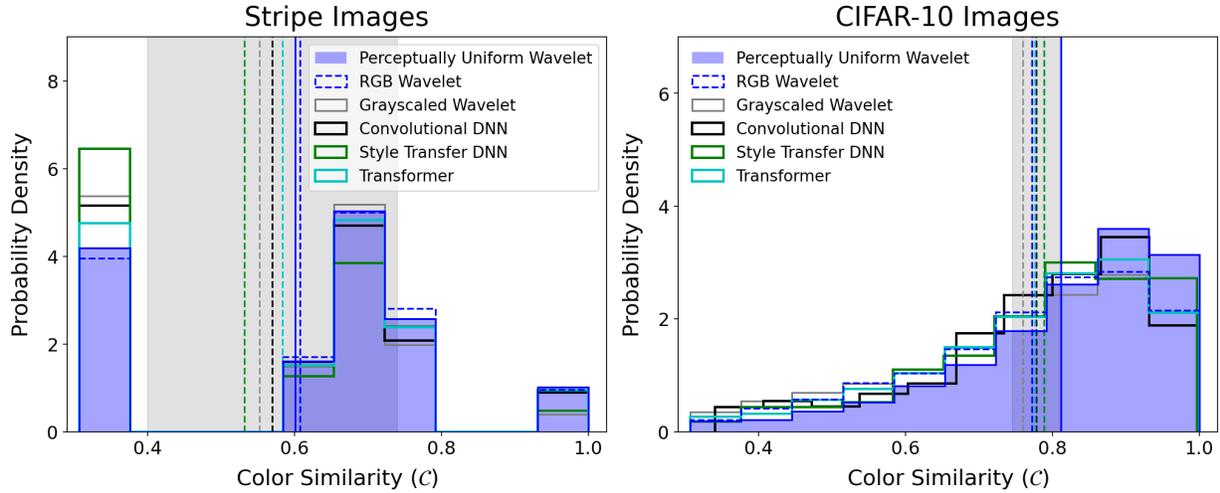

**Figure S16.** Color similarity distributions for our stripe (left) and CIFAR-10 (right) image datasets returned by our perceptually uniform wavelet (blue), RGB wavelet (unfilled blue), grayscaled wavelet (gray), convolutional DNN (black), style transfer DNN (green), and vision transformer (cyan) algorithms. For both datasets, our perceptually uniform wavelet algorithm yields image clusters with the highest mean color similarity.

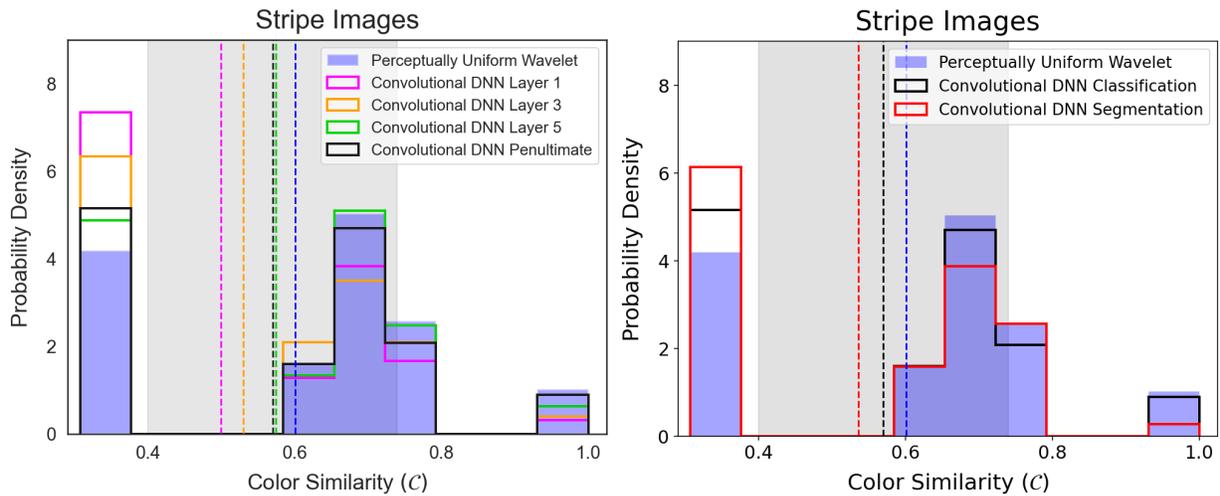

**Figure S17.** Left panel: stripe dataset color similarity distributions for our perceptually uniform wavelet algorithm (blue) and our convolutional DNN using the first (magenta), third (orange), fifth (green), and penultimate (black) embedding layers. Right panel: same as the left panel, but comparing our wavelet algorithm to our fiducial convolutional DNN trained on image classification using the ResNet architecture (black) versus a convolutional DNN trained on image segmentation using the U-Net architecture (red).

*Robustness to Image Resolution*

We also demonstrate that our results are robust to image resolution. In particular, we recreate our color similarity analysis by embedding a higher-resolution version of the same CIFAR-10 dataset (128 x 128 pixels, rather than our standard 32 x 32 pixels) with our convolutional DNN. Figure S18 shows that the resulting color similarity distributions are virtually identical to our fiducial results, indicating that image resolution does not affect our conclusions.

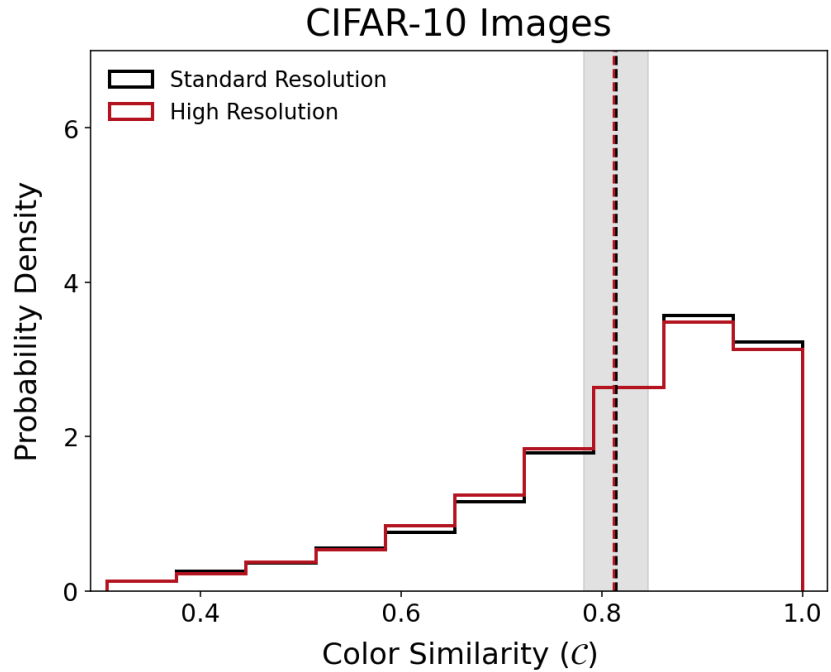

**Figure S18.** Color similarity distributions for the CIFAR-10 clusters returned by our convolutional DNN algorithm using our standard (32 x 32)-pixel resolution CIFAR-10 images (black) and a higher-resolution (128 x 128)-pixel versions of the same images (brown). The distributions are virtually identical; thus, our results are not sensitive to image resolution.

**Study 1: D. A Color Vision Test for Computer Vision Algorithms**

*Embedding Similarity versus Luminance*

It is interesting to consider how specific properties of image pairs' color distributions correlate with embedding similarity. Thus, Figure S19 shows the relationship between embedding similarity and the mean $J_z$ coordinate of each block pair, where the mean $J_z$ coordinates are minmax-normalized. Intriguingly, our wavelet algorithm's response to image pair luminance is much more symmetric than for any of the DNNs we test. For example, the probability that the convolutional DNN embeds block image pairs with low luminance (i.e., mean pairwise $J_z < 0.5$ in Figure S19) less similarly than average is more than twice that for our wavelet algorithm. This suggests that luminance drives the convolutional DNN's embeddings of images in different parts of color space, such that it represents low-luminance colors similarly regardless of their perceptual differences in hue and saturation.

Figure S19 indicates that similar conclusions hold for the style transfer DNN, which again behaves very similarly to the convolutional DNN. Meanwhile, the vision transformer DNN tends to embed all block images similarly, regardless of their luminance. This is consistent with our results in Figure 5 and again suggests that the uniform texture of our block images exacerbates the non-perceptual behavior of the vision transformer DNN we test.

The non-perceptual responses of convolutional and style transfer DNNs to low-luminance images is reminiscent of known shortcomings of RGB color space, in which low-luminance colors are less well differentiated than high-luminance colors[5]. Thus, the DNN behavior we identify in Figure S19 could be caused by the fact that these algorithms are trained on images represented in non-perceptually uniform color spaces. However, these findings may also signal inherent biases in these algorithms' image embeddings, regardless of the color space used when representing images during training.

*Robustness to Block Image Color Contrast*

In our standard block image dataset, each color tile is laid on a white background. Here, we test two alternative background colors to show that color contrast does not affect our conclusions. In particular, we generate two alternative versions of the block dataset with identical central colors but with black and gray backgrounds, respectively, and replicate our analyses using the convolutional DNN. In these tests, color similarity between each block image pair is unchanged because the border is not used when computing color similarity; however, embedding similarity changes because the entire image is embedded by the convolutional DNN.

As shown in Figure S20, these background color variations do not qualitatively affect our findings: the correlation between color similarity and embedding similarity for the convolutional DNNs is significantly lower than for the wavelet algorithm ($\rho = 0.95$) when using either a white ($\rho = 0.5$, $p < 0.001$), black ($\rho = 0.53$, $p < 0.001$), or gray ($\rho = 0.6$, $p < 0.001$) background (Student t-test). Furthermore, the overall shape of the embedding similarity–color similarity relation is qualitatively unchanged across these different background conditions. Thus, we conclude that block image backgrounds and color contrast do not affect our main findings.

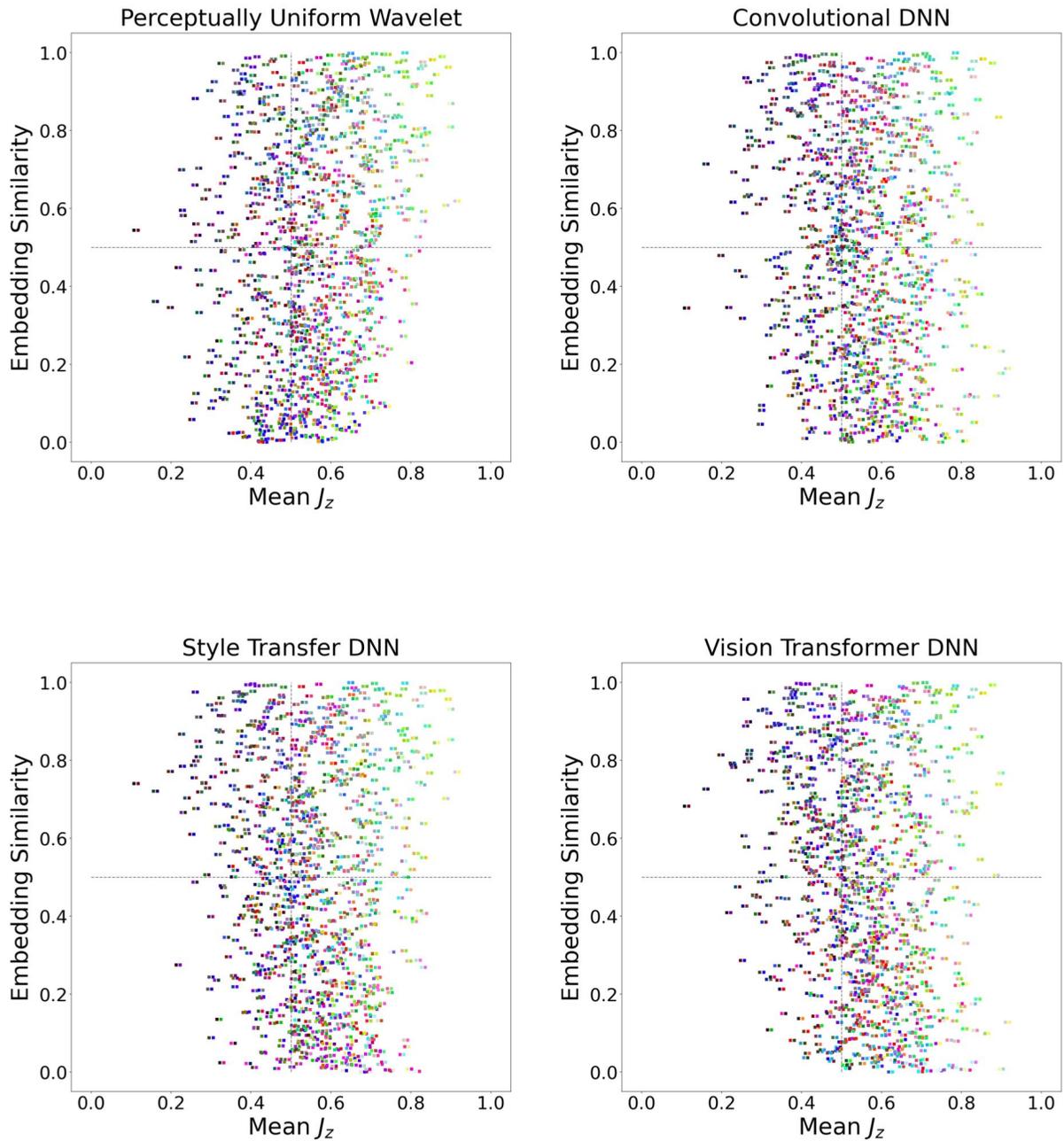

**Figure S19.** Perceptually uniform wavelet (top left), convolutional DNN (top right), style transfer DNN (bottom left), and vision transformer DNN (bottom right) embedding similarity versus the mean $J_z$ coordinate for a fixed random sample of block image pairs. Our wavelet algorithm responds to low and high-luminance images more symmetrically than any of the DNNs, which tend to embed low-luminance images similarly, regardless of hue and saturation.

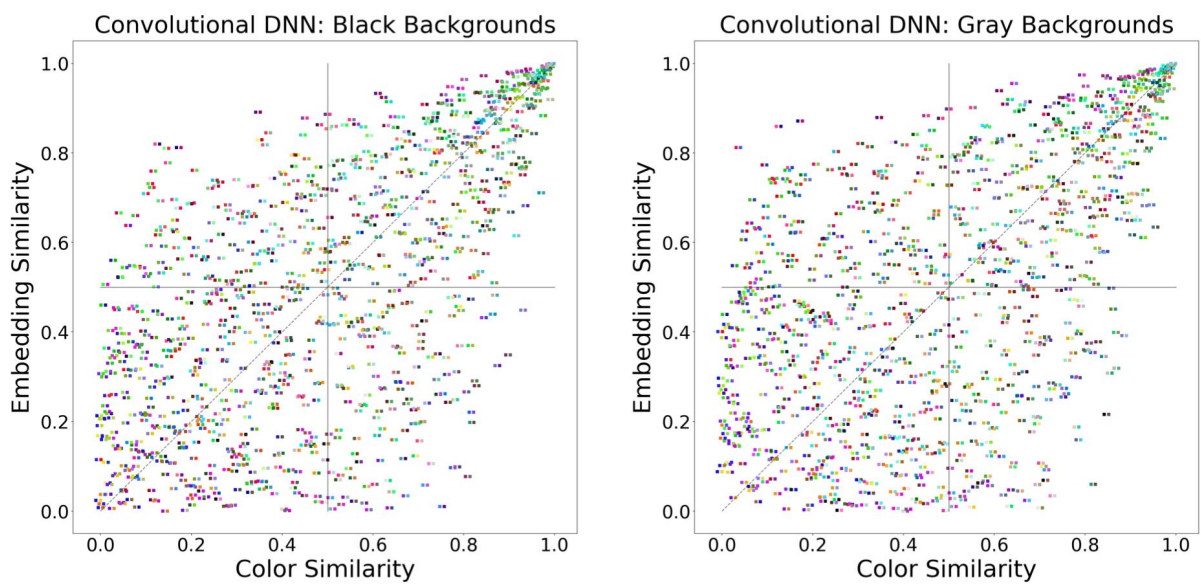

**Figure S20.** Convolutional DNN embedding similarity versus color similarity for alternative versions of our block image dataset with black (left) and gray (right) backgrounds, rather than our standard choice of white backgrounds. The embedding similarity–color similarity relation is not qualitatively changed by varying image background and color contrast, and the corresponding correlation coefficients are also not significantly affected. Each panel shows an independent sample of 1000 random block image pairs.

**Study 2: A. Predicting Human Color Judgments from an Online Survey**

*Perceptual Errors in Algorithms' Color Judgments*

Figure S21 shows embedding similarity from our wavelet and convolutional, style transfer, and vision transformer DNN algorithms versus perceptually uniform color similarity for the color tile pairs included in our survey. The axes are analogous to Figure 5, except that the embedding and color similarities are not rank-ordered before minmax normalization is applied, to better highlight color tile pairs that are outliers in either dimension. Lines connect pairs for which a given algorithm failed to provide a color similarity judgment consistent with the human consensus. For example, the convolutional DNN connects the pair of light green and light yellow tiles in the bottom-right quadrant of the right-hand panel with the purple and light brown pair in the bottom-left panel. Humans judge the green/yellow pair as more similar, while the convolutional DNN embeds the purple/brown pair more similarly.

Again, the style transfer DNN behaves similarly to the convolutional DNN; for example, it yields several perceptual errors that involve the same purple/brown color tile pair described above. In addition, many of the perceptual errors of both the convolutional and style transfer DNNs involve pairs of primary colors (e.g., see the lines connecting to the red/blue pairs at the top-left of each algorithm's panel in Figure S21). Finally, as discussed above, the vision transformer DNN embeds all block images similarly. It is therefore difficult to pinpoint regions of color space in which it fails to represent human judgments, but we note that several of its perceptual errors involve the outlying color tile pairs that it embeds differently very differently than the remaining pairs (for example, see the pairs that contain a cyan tile towards the bottom-right of the vision transformer DNN panel in Figure S21).

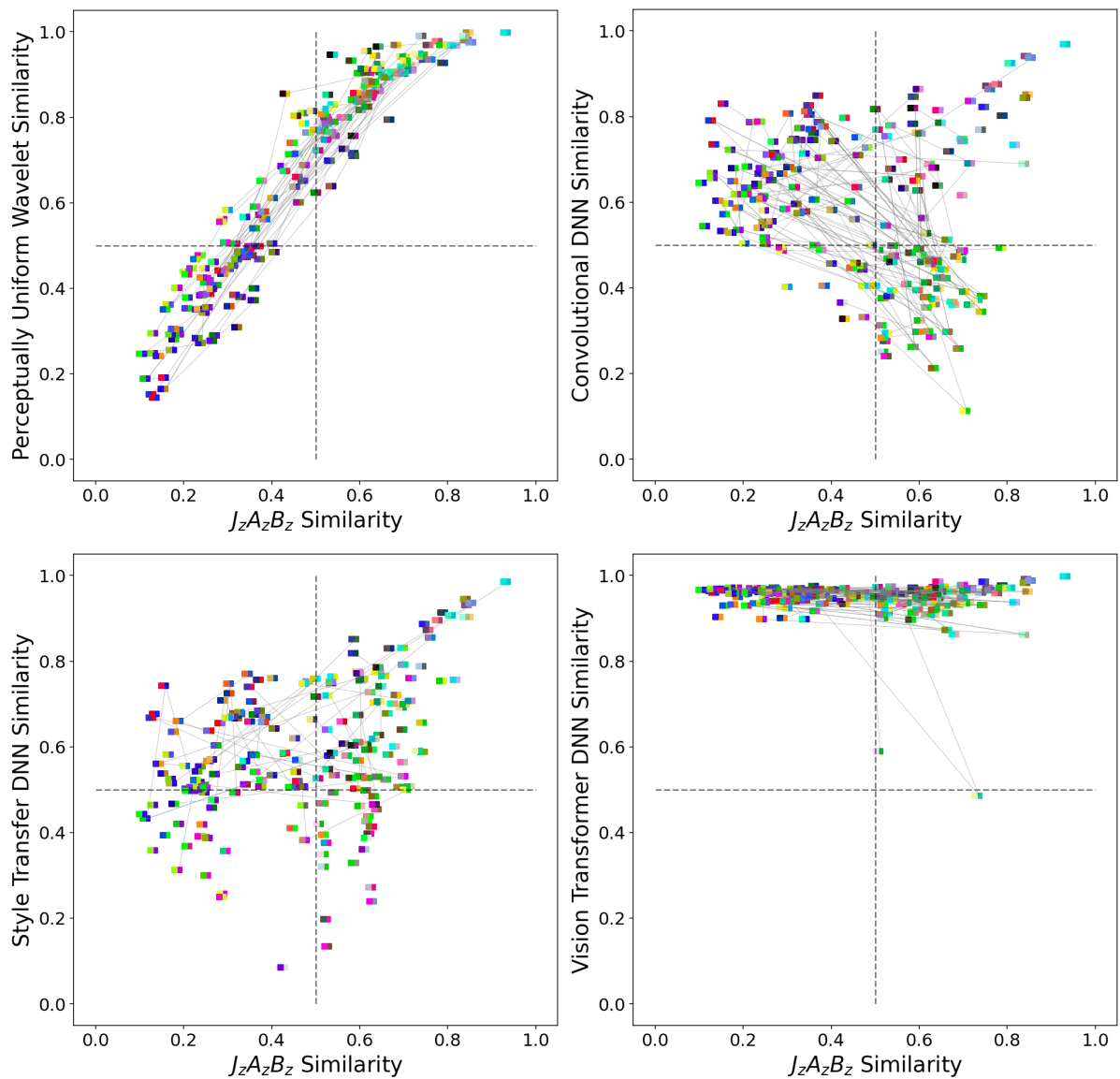

**Figure S21.** Perceptually uniform wavelet (top left), convolutional DNN (top right), style transfer DNN (bottom left), and vision transformer DNN (bottom right) embedding similarity versus the $J_zA_zB_z$ similarity of each image pair used in our online survey. Lines connect pairs for which a given algorithm failed to provide a color similarity judgment consistent with the human consensus.

# Supplementary References